\documentclass[10pt,twocolumn,letterpaper]{article}

\usepackage[pagenumbers]{wacv} % To force page numbers, e.g. for an arXiv version

\definecolor{wacvblue}{rgb}{0.21,0.49,0.74}
\usepackage[pagebackref,breaklinks,colorlinks,allcolors=wacvblue]{hyperref}
\usepackage{graphicx}
\usepackage{booktabs}
\usepackage{caption}
\usepackage{subcaption}
\usepackage{amsmath,amssymb}
\usepackage{multirow}
\usepackage{makecell}
\usepackage{adjustbox}
\usepackage{float}
\usepackage{color}
\usepackage{xcolor, colortbl}
\usepackage{pifont}
\usepackage{wrapfig}

\usepackage{algorithm}
\usepackage{algpseudocode}
\algrenewcommand\algorithmicindent{1.8em}

\newcommand{\paratitle}[1]{\noindent\textbf{#1}\space\ignorespaces}

\def\wacvPaperID{856} % *** Enter the WACV Paper ID here
\def\confName{WACV}
\def\confYear{2027}

\title{Embedding Predictive Adaptation and Refinement for Multi-Modal Industrial Anomaly Detection}

\author{Jaeho Cho \quad Jinsik Bang \quad Taehwan Kim\\
Ulsan National Institute of Science and Technology (UNIST), Republic of Korea\\
{\tt\small \{jh\_cho, bang, taehwankim\}@unist.ac.kr}
}

\AtBeginDocument{%
  \setlength{\abovedisplayskip}{2pt}%
  \setlength{\belowdisplayskip}{2pt}%
  \setlength{\abovedisplayshortskip}{2pt}%
  \setlength{\belowdisplayshortskip}{2pt}%
}

\begin{document}
\maketitle
% \thispagestyle{empty}
% \vspace{-2em}
% \enlargethispage{\baselineskip}     
\begin{abstract}
Industrial Anomaly Detection (IAD) has evolved from 2D image-based approaches to multi-modal methods that leverage the complementary nature of RGB images and 3D point clouds. Among these, memory bank-based methods, which identify anomalies by the discrepancy between each query patch feature and its nearest normal prototype, have proven particularly effective. However, two limitations remain: (i) employing frozen pretrained encoders yields insufficient discrimination between normal and anomalous features due to the domain gap, while existing contrastive learning-based adaptation struggles to capture the spatial relationships required for precise localization; and (ii) pixel-level anomaly maps interpolated from patch-level distance scores remain coarse, lacking fine-grained detail. To address these limitations, we propose EPAR (Embedding Predictive Adaptation and Refinement). We adapt encoders by predicting the embeddings of masked regions to capture spatial relationships, together with patch-level Variance-Invariance-Covariance (VIC) regularization, resolving the poor discriminability between normal and anomalous features caused by the domain gap. We further propose a Noise Suppression Network that refines coarse maps by learning the relational discrepancy between each patch feature and its nearest normal prototype, rather than the anomaly's appearance. This relational signal remains valid for unseen real defects, despite being learned from synthetic anomalies. EPAR demonstrates strong practical applicability, excelling in few-shot settings and improving AUPRO@1\% on MVTec 3D-AD and Eyecandies by +1.8\% and +1.9\%, respectively. Source code is available at \url{https://github.com/ohcoheaj/EPAR}.
\end{abstract}    
\begin{figure}[t]
\centering
\includegraphics[width=1.0\linewidth]{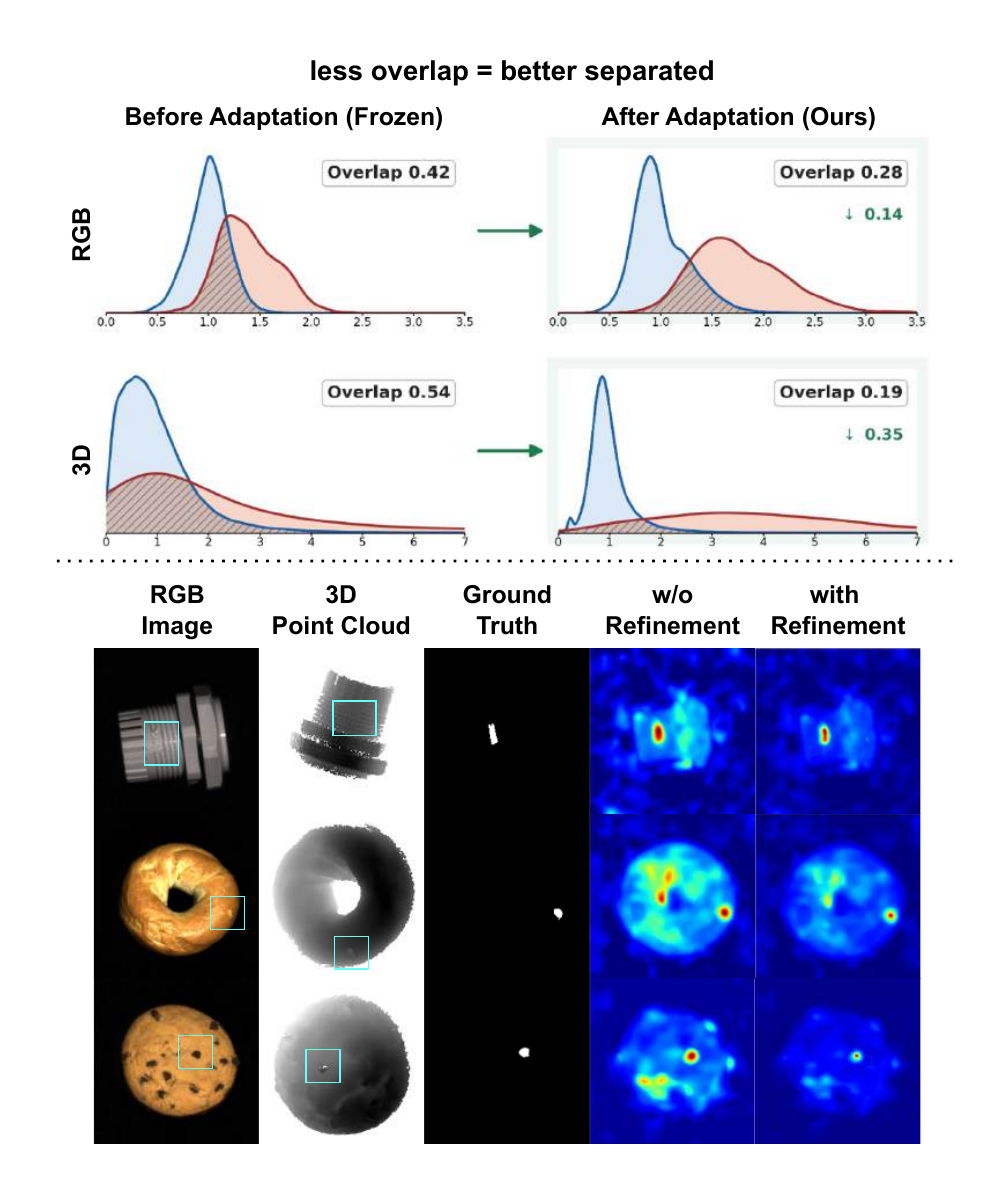}
\vspace{-3em}
\caption{\textbf{EPAR at a glance.}
EPAR comprises a feature adaptation stage and a refinement stage. \textbf{(top)} For each query patch feature, we measure the distance to its nearest normal prototype in the memory bank, shown as densities normalized to unit area for normal (\textcolor{blue}{blue curve}) and anomalous (\textcolor{red}{red curve}) patches. The shaded region marks their overlap, quantified by the overlap coefficient (lower is better). After adaptation, the overlap narrows for both RGB and 3D, \ie, normal and anomalous patches become better separated. \textbf{(bottom)} Building on the adapted features, our refinement sharpens coarse anomaly maps, suppressing false positives while preserving true anomalies.}
\label{fig:teaser}
\end{figure}

\vspace{-1.8em} 
\section{Introduction}
% \enlargethispage{\baselineskip}
% What is IAD?
\vspace{-0.5em}
Industrial Anomaly Detection (IAD) is critical for product quality assurance, comprising sample-level detection that determines whether a sample contains defects and pixel-level segmentation of defective regions. 
Real-world defects are rare and unpredictable, making supervised approaches~\cite{das,unifyinganalysis} impractical. Labeling defects is costly and such approaches struggle to generalize to unseen anomaly types.
To overcome this, an unsupervised paradigm~\cite{patchcore,padim,revdistill} trains exclusively on nominal samples and quantifies discrepancies at inference.
% Flow (2D -> 2D + 3D)
While 2D RGB image-based approaches~\cite{patchcore,padim,unistudent,d&a,SPCAB,revdistill,cflowad} have been extensively studied on datasets such as MVTec AD~\cite{mvtecad}, real-world manufacturing defects often involve geometric irregularities that are difficult to detect from RGB images alone. 
To address this, recent studies~\cite{M3DM,LSFA,CFM,g2sf} have explored multi-modal frameworks that leverage the complementary nature of RGB and 3D point clouds to detect diverse defect types on datasets such as MVTec 3D-AD~\cite{mvtec3d} and Eyecandies~\cite{eyecandies}.

% Representative Method
In both 2D and multi-modal settings, feature embedding-based methods~\cite{patchcore,tocoad,reconfa,recontrast,M3DM,LSFA,g2sf,CFM} leverage encoders pretrained on large-scale datasets (e.g., ImageNet~\cite{imagenet} for RGB, ShapeNet~\cite{shapenet} for 3D) to extract semantically meaningful features. 
Among these, memory bank-based methods~\cite{patchcore,M3DM,LSFA,g2sf,tocoad,reconfa} have proven particularly effective.
They store representative patch-level features (\ie, normal prototypes) from nominal samples and, at inference, identify anomalies by the discrepancy between each query patch feature and its nearest normal prototype.
They differ in how they utilize pretrained encoders: (i) using frozen encoders as feature extractors~\cite{patchcore,M3DM,g2sf} and (ii) adapting features to the target domain~\cite{tocoad,reconfa,LSFA}. 
While the former is widely adopted, the domain gap between source and target data leads to insufficient discrimination between normal and anomalous features, potentially causing false alarms and missed detections~\cite{tocoad,LSFA} (Fig.~\ref{fig:teaser}). 
% Limitation of representative method
Within multi-modal settings, which are the focus of this work, LSFA~\cite{LSFA} addresses the domain gap via contrastive learning-based feature adaptation. 
While effective for sample-level detection, contrastive learning prioritizes invariance between positive pairs over spatial relationships among neighboring regions~\cite{cmae,maskvscont,pqcl}, and thus may not fully exploit the fine-grained spatial information required for precise pixel-level segmentation~\cite{fcl}.

To capture such spatial relationships, we build on mask prediction-based feature adaptation~\cite{enhancingjepa,maeanal}, which predicts masked regions from surrounding context. 
There are two main strategies: Reconstruction-based~\cite{MAE,pointmae} and Embedding prediction-based~\cite{ijepa,pointjepa}. 
We adopt the latter, since embedding prediction captures what each region semantically represents within its context rather than low-level appearance. 
This is well-suited for anomaly detection, where a defect is readily identified when a local patch's feature deviates from what its surrounding context implies~\cite{glancing}.
However, as a joint-embedding architecture, embedding prediction alone is prone to representation collapse~\cite{vicreg,cjepa}. This is detrimental to IAD, which relies on discrepancy between normal and anomalous features. 
To prevent collapse, C-JEPA~\cite{cjepa} combines embedding prediction with Variance-Invariance-Covariance (VIC) regularization~\cite{vicreg}, ensuring each dimension encodes distinct information. We adopt this principle, tailored to memory bank-based IAD.
As anomalies are identified by per-patch feature discrepancy, we apply VIC regularization across patches within each sample rather than across a batch~\cite{vicreg,cjepa}. We observe that this patch-level formulation outperforms the standard batch-level one (Sec.~\ref{subsec:ablation}).

Building upon feature adaptation, we address a limitation of pixel-level anomaly maps of existing methods~\cite{M3DM,LSFA,g2sf}. Current memory bank-based methods obtain pixel-level maps by interpolating patch-level distance scores. This procedure yields coarse maps and can highlight nominal regions as anomalous. Such false positives undermine the reliability of defect localization.
To address this, we propose a Noise Suppression Network (NSN), a refinement strategy applicable to both RGB and 3D modalities. Prior refinement methods~\cite{pni,reconpatch} learn to directly refine the coarse maps of synthesized samples under pseudo-mask supervision to improve the reliability of defect localization. Since synthetic anomalies are explicitly distinct in appearance from real defects, this risks learning shortcuts to synthesis-specific cues. NSN instead operates in feature space, pairing each patch feature with its nearest normal prototype and learning their discrepancy rather than the anomaly's appearance.
As this relational signal is independent of how an anomaly looks, we observe that NSN generalizes from pseudo-anomalies at training to unseen real anomalies at inference, suppressing false positives while preserving true anomalies~(Fig.~\ref{fig:qualitative_ablation}).

Our contributions are summarized as follows:
\begin{itemize}
\item[$\bullet$] We introduce Embedding Predictive Adaptation, a feature adaptation scheme tailored to memory bank-based IAD. It captures spatial relationships via embedding prediction, while patch-level VIC regularization prevents collapse and yields distinct features that benefit the nearest-neighbor scoring of memory banks.
% \item[$\bullet$] We propose a Noise Suppression Network (NSN) that learns the relational discrepancy between each patch feature and its nearest normal prototype rather than the appearance of synthetic anomalies. This relational signal remains valid for unseen real defects, despite being learned from synthetic anomalies alone.
\item[$\bullet$] We propose a Noise Suppression Network (NSN) that learns the relational discrepancy between each patch feature and its nearest normal prototype rather than the appearance of synthetic anomalies. This relational signal remains valid for unseen real defects, enabling 
precise and reliable defect localization.
% \item[$\bullet$] EPAR achieves state-of-the-art performance in I-AUROC and AUPRO@1\% on MVTec 3D-AD~\cite{mvtec3d} and outperforms all baselines across all metrics on Eyecandies~\cite{eyecandies}.
\item[$\bullet$] EPAR achieves state-of-the-art performance in I-AUROC and AUPRO@1\% on MVTec 3D-AD~\cite{mvtec3d} and Eyecandies~\cite{eyecandies}. EPAR exhibits strong practical applicability, supported by superior few-shot performance and consistent gains under AUPRO@1\%, a metric reflecting stringent false-alarm requirements for real-world deployment.
\end{itemize}
% \begin{figure*}[t]
%     \centering
%     \includegraphics[width=1.0\linewidth]{1st_stage_1column.pdf}
%     \vspace{-2.5em}
%     \caption{\textbf{Feature Adaptation Stage.}
%     \textbf{(a) Input Preparation.} For 3D point clouds, patches are formed via Farthest Point Sampling (FPS)~\cite{fps} (1024 centers) and KNN (128 points per patch). Target blocks (distinct colors) are sampled from foreground, with the remaining regions forming the context input. 
%     \textbf{(b) Adaptation Process.} The context encoder extracts features $C_k$ from visible regions at layers $k \in \{\text{low}, \text{mid}, \text{high}\}$, while the target encoder processes the complete input to provide target representations. Layer-specific predictors $\Phi_k$ predict masked regions from context features with learnable mask tokens, enabling multi-scale adaptation.
%     \textbf{(c) VIC Regularization.} We integrate VIC regularization~\cite{vicreg} at the patch-level with mask prediction: the variance term maintains feature diversity, the invariance term enforces $c_i$-$t_i$ consistency, and the covariance term decorrelates feature dimensions.}
%     \label{fig:jepa_figure}
% \end{figure*}

\vspace{-0.5em}
\section{Related Work}
\subsection{Multi-Modal Industrial Anomaly Detection}
Multi-modal industrial anomaly detection aims to detect diverse defect types by leveraging the complementary nature of RGB and 3D.
On multi-modal datasets such as MVTec 3D-AD~\cite{mvtec3d} and Eyecandies~\cite{eyecandies}, most recent methods~\cite{M3DM,LSFA,g2sf} follow the successful 2D paradigm~\cite{patchcore}: they store representative patch-level features (\ie, normal prototypes) extracted from nominal samples and identify anomalies by the discrepancy between each query patch feature and its nearest prototype via Nearest Neighbor Search (NNS).
These methods fall into two groups according to how they utilize pretrained encoders (\eg, DINO~\cite{dino} for RGB, PointMAE~\cite{pointmae} for 3D).
The first group keeps encoders frozen. M3DM~\cite{M3DM} stores RGB, 3D, and fused features in separate banks, followed by aggregation of their outputs via learnable OCSVMs~\cite{ocsvm}. G$^2$SF~\cite{g2sf} learns an anisotropic distance metric over per-modality banks using synthetic anomalies.
However, frozen encoders leave the domain gap between source and target data unaddressed.
The second group adapts features to the target domain: LSFA~\cite{LSFA} fine-tunes lightweight adapters via intra-modal compactness optimization and cross-modal local-to-global contrastive alignment, averaging per-modality anomaly maps at inference.
Departing from memory banks, CFM~\cite{CFM} maps features across modalities with lightweight MLPs and scores anomalies by the discrepancy between observed and mapped features.
We also build upon the memory bank paradigm given its strong performance, but extend it in two directions: embedding prediction-based feature adaptation to bridge the domain gap, and a refinement stage that sharpens the interpolation-based anomaly maps before fusion.

\vspace{-0.3em}
\subsection{Embedding Prediction-based Self-Supervision}
\vspace{-0.3em}
Mask prediction-based approaches capture spatial relationships by predicting masked regions from surrounding context~\cite{maeanal,enhancingjepa}.
Reconstruction-based approaches~\cite{MAE,pointmae,pilamim} reconstruct randomly masked regions in the input space, whereas embedding prediction-based approaches~\cite{ijepa,pointjepa} predict the latent embeddings of the spatially contiguous masked regions.
% Among embedding prediction-based approaches, C-JEPA~\cite{cjepa} integrates VIC regularization~\cite{vicreg} to prevent the representation collapse inherent in joint-embedding architectures, improving both global recognition and dense prediction in general-purpose pretraining.
Among the latter, C-JEPA~\cite{cjepa} integrates VIC regularization~\cite{vicreg} to prevent the representation collapse inherent in joint-embedding architectures.
% We employ this principle not for general-purpose pretraining but for adapting frozen multi-modal encoders to the target industrial domain.
% Moreover, since memory bank-based detection identifies anomalies based on per-patch feature discrepancy, we reformulate VIC regularization across patches within each sample, rather than across samples within a batch as in~\cite{vicreg,cjepa}.
We employ this principle instead for adapting frozen encoders to the target industrial domain. 
Moreover, since memory bank-based detection identifies anomalies by per-patch feature discrepancy, we reformulate VIC regularization across patches within each sample rather than across samples within a batch~\cite{vicreg,cjepa}.

\vspace{-0.3em}
\subsection{Anomaly Map Refinement}
\vspace{-0.3em}
Pixel-level anomaly maps interpolated from patch-level scores often lack fine-grained detail~\cite{pni,reconpatch}. To sharpen them, PNI~\cite{pni} and ReConPatch~\cite{reconpatch} synthesize pseudo-anomalies and train a refinement network under pseudo-mask supervision. Such appearance-driven supervision, however, risks learning shortcuts specific to synthetic anomalies. Our NSN instead operates in the feature space, learning the relational discrepancy between each patch feature and its nearest normal prototype, which remains valid for unseen real defects.
\begin{figure*}[t]
    \centering
    \includegraphics[width=1.0\linewidth]{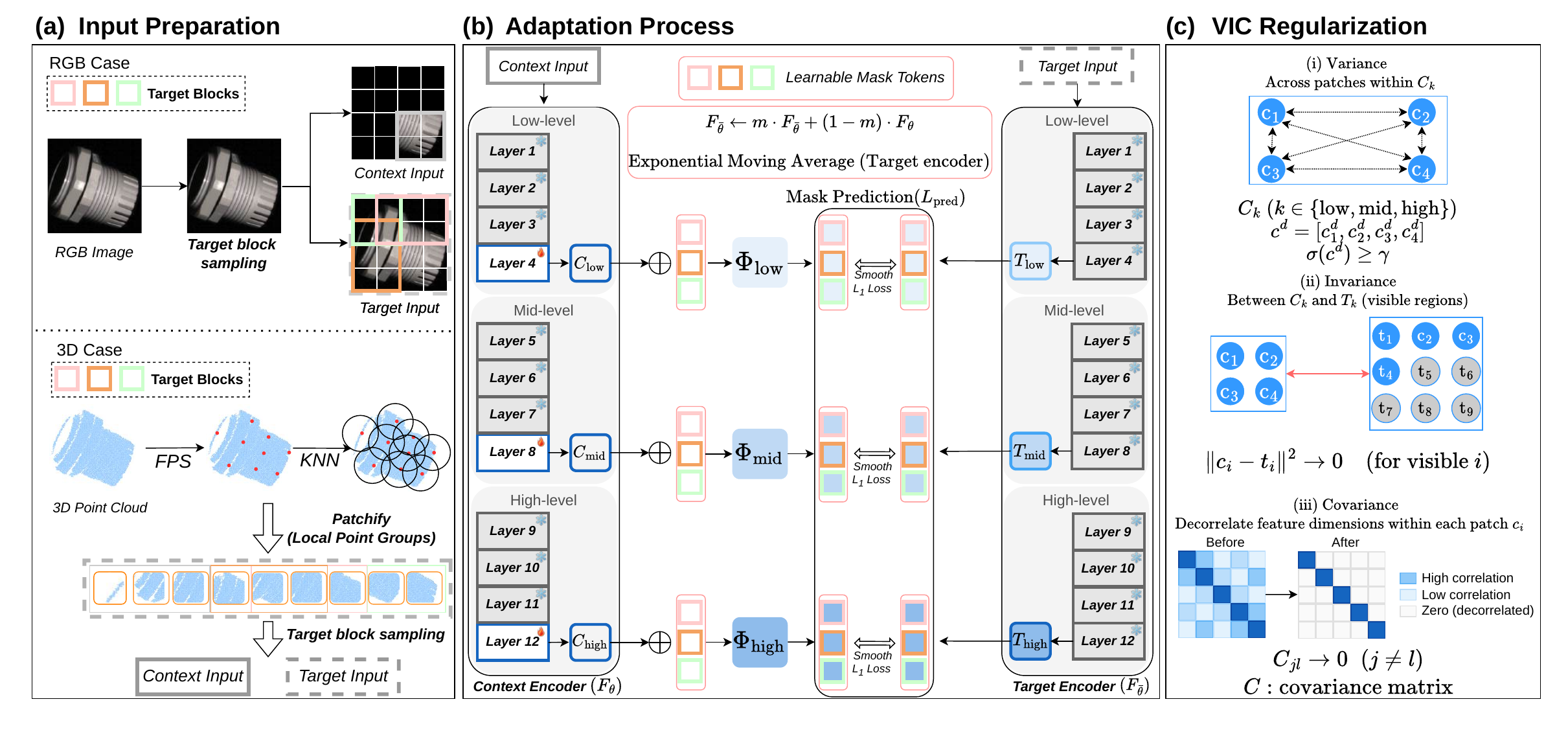}
    \vspace{-2.5em}
    \caption{\textbf{Feature Adaptation Stage.}
    \textbf{(a) Input Preparation.} For 3D point clouds, patches are formed via Farthest Point Sampling (FPS)~\cite{fps} (1024 centers) and KNN (128 points per patch). Target blocks (distinct colors) are sampled from foreground, with the remaining regions forming the context input. 
    \textbf{(b) Adaptation Process.} The context encoder extracts features $C_k$ from visible regions at layers $k \in \{\text{low}, \text{mid}, \text{high}\}$, while the target encoder processes the complete input to provide target representations. Layer-specific predictors $\Phi_k$ predict masked regions from context features with learnable mask tokens, enabling multi-scale adaptation.
    \textbf{(c) VIC Regularization.} We integrate VIC regularization~\cite{vicreg} at the patch-level with mask prediction: the variance term maintains feature diversity, the invariance term enforces $c_i$-$t_i$ consistency, and the covariance term decorrelates feature dimensions.}
    \label{fig:jepa_figure}
\end{figure*}

% \vspace{-0.2em}
\section{Method}
\label{sec:Method}
\vspace{-0.3em}

Across multi-modal settings~\cite{M3DM,LSFA,g2sf}, the final anomaly prediction is produced by combining spatially-aligned per-modality maps to exploit the complementary nature of RGB and 3D. Since the fused result is highly dependent on the maps being combined, we aim to produce precise per-modality maps through two stages: feature adaptation that bridges the domain gap, and refinement that sharpens interpolation-based coarse maps into fine-grained ones.

\subsection{Feature Adaptation}
\vspace{-0.1em}
\label{subsec:feature_adapt}
\noindent\textbf{Embedding Prediction.} 
As illustrated in Fig.~\ref{fig:jepa_figure}, we adapt pretrained features using only normal samples by predicting the embeddings of spatially contiguous masked blocks~\cite{ijepa,cjepa,pointjepa}.
Following JEPA~\cite{ijepa,pointjepa}, the adaptation employs a dual-encoder structure: a context encoder ($F_{\theta}$) encodes the visible regions (\ie, excluding the masked regions) into context features, whereas a target encoder ($F_{\bar{\theta}}$), updated via exponential moving average of the context encoder, encodes the full input into target embeddings.
A lightweight ViT-based predictor~\cite{vit} then takes the context features combined with position-aware learnable mask tokens as input and predicts the target embeddings of spatially contiguous masked blocks (4 per sample~\cite{ijepa,pointjepa}). 
Both encoders are initialized with the same modality-specific pretrained weights (DINO~\cite{dino} for RGB, PointMAE~\cite{pointmae} for 3D).
These target blocks are selected only from foreground regions to capture object semantics rather than the background.
We obtain the foreground from the 3D data. Since each RGB pixel corresponds to a 3D point at the same location, we remove the background plane from the point cloud via RANSAC~\cite{RANSAC} and apply the same foreground region to the RGB image, following~\cite{M3DM,BTF}.

\noindent\textbf{Multi-Scale Feature Extraction.}
Prior works~\cite{M3DM,LSFA,g2sf} concatenate features from multiple intermediate layers (4/8/12th) for 3D, yet use only the final-layer features for RGB. Since features at different depths encode qualitatively distinct information~\cite{dinoanal}, we adapt multiple intermediate layers for both modalities, employing a separate predictor at each layer for layer-specific adaptation.

\noindent\textbf{Training Objective.} Our adaptation objective combines embedding prediction with VIC regularization~\cite{vicreg}.
Let $\hat{z}_j$ denote the predicted representation from the predictor for masked region $j$, $z_j$ denote the corresponding target encoder output, and $M$ denote the total number of masked patches across all target blocks. The embedding prediction loss $\mathcal{L}_{\text{pred}}$ is the smooth $\ell_1$ loss between the predicted and target representations for masked regions:
\begin{equation} 
\mathcal{L}_{\text{pred}} = \frac{1}{M}\sum_{j=1}^{M} 
\begin{cases} \frac{1}{2}(\hat{z}_j - z_j)^2 & \text{if } |\hat{z}_j - z_j| < 1 \\ 
|\hat{z}_j - z_j| - \frac{1}{2} & \text{otherwise}
\end{cases} 
\end{equation} 
For VIC regularization, let $c_i$ and $t_i$ denote the context and target encoder outputs for visible region $i$. The context features at layer $k\in\{\text{low},\text{mid},\text{high}\}$ are stacked as $C_k=[c_1,\ldots,c_N]$, where $N$ is the number of visible patches and $D$ the feature dimension. $\mathcal{L}_{\text{var}}$ maintains the variance of each dimension across patches, $\mathcal{L}_{\text{cov}}$ reduces the covariance between dimensions, and $\mathcal{L}_{\text{inv}}$ enforces consistency between context and target outputs:
\begin{equation}
\resizebox{\columnwidth}{!}{$
\displaystyle
\mathcal{L}_{\text{var}} = \frac{1}{D}\sum_{d=1}^{D} \max(0,\,\gamma - \sigma_d),
\quad
\mathcal{L}_{\text{inv}} = \frac{1}{N}\sum_{i=1}^{N} \|c_i - t_i\|_2^2,
\quad
\mathcal{L}_{\text{cov}} = \frac{1}{D}\sum_{j \neq l} C_{jl}^2
$}
\end{equation}
where $\sigma_d = \sqrt{\text{Var}(c^d) + \epsilon}$ is computed over the $d$-th dimension across patches, $\gamma$ is the target standard deviation, and $C$ is the covariance matrix over patches in $C_k$.
The overall adaptation objective is:
\begin{equation} 
\mathcal{L}_{\text{adapt}} = \mathcal{L}_{\text{pred}} + \alpha \cdot 
(\lambda_{\text{var}}\mathcal{L}_{\text{var}} + \lambda_{\text{inv}}\mathcal{L}_{\text{inv}} + \lambda_{\text{cov}}\mathcal{L}_{\text{cov}}) 
\end{equation}

\begin{figure*}[t]
    \centering
    \includegraphics[width=0.95\textwidth]{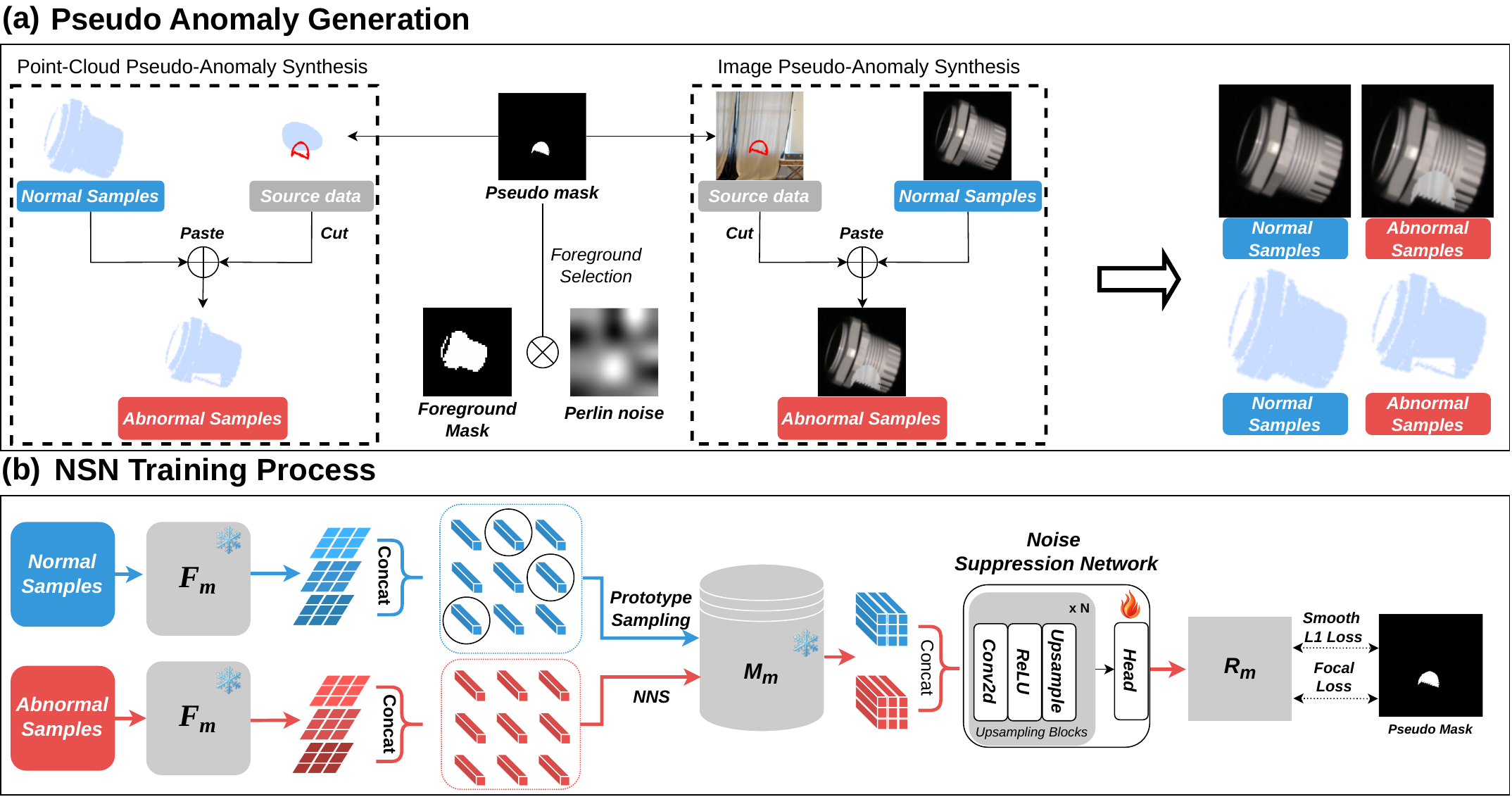}
    \vspace{-0.5em}
    \caption{\textbf{Noise Suppression Network Training Process.} \textbf{(a)} illustrates the process of synthesizing pseudo-anomalies. \textbf{(b)} illustrates the NSN training process for each modality $m \in \{\text{RGB}, \text{3D}\}$. The \textcolor{blue}{blue path} shows constructing the memory bank $\mathcal{M}_m$ via prototype sampling~\cite{patchcore} from normal features. The \textcolor{red}{red path} shows processing pseudo-anomalies through the adapted encoders. Each patch feature $f_i$ is paired with its nearest neighbor via Nearest Neighbor Search (NNS) $f^*_i$ from $\mathcal{M}_m$ to form $[f_i, f^*_i]$ for supervised NSN training.}
    \label{fig:2nd_stage_figure}
\end{figure*}

\subsection{Training Noise Suppression Network}
\vspace{-0.1em}
\label{subsec:nsn}
As illustrated in Fig.~\ref{fig:2nd_stage_figure}, we propose a Noise Suppression Network (NSN) that refines coarse anomaly maps by learning the relationship between each query patch feature and its nearest normal prototype. This relational signal enables the network to learn the discrepancy between normal and anomalous features rather than the appearance of defects.

\noindent\textbf{Pseudo-Anomaly Synthesis.} We generate pseudo-anomalies by a cut-and-paste strategy. Perlin noise is thresholded into a binary mask, which is then restricted to the foreground region obtained in Sec.~\ref{subsec:feature_adapt} so that anomalies are placed only on the object, yielding the pseudo-mask $M^*$. We then cut the region defined by $M^*$ from an anomaly source and paste it onto the normal sample. For RGB, the anomaly source is randomly augmented DTD~\cite{dtd} textures, following~\cite{reconpatch,memseg,prototypical}. For 3D, the anomaly source is randomly transformed samples of other categories, following~\cite{g2sf}.

\noindent{\textbf{Architecture and Training.}}
We first construct compact memory banks for each modality through prototype sampling following previous works~\cite{M3DM,g2sf,patchcore}. For 3D features, we project the group-level features onto a 2D grid via spatial alignment~\cite{M3DM,LSFA} before constructing the memory bank, placing them on the same spatial layout as the RGB features for the subsequent late fusion. To train the NSN, we extract features from pseudo-anomalies using the frozen adapted encoders, and for each patch feature $f_i$ we retrieve its nearest normal prototype $f^*_i$ from the memory bank based on $\ell_2$ distance. The concatenated pair $[f_i, f^*_i] \in \mathbb{R}^{2D}$ is fed into the NSN, which comprises CNN-based upsampling blocks followed by a segmentation head, producing a pixel-level refinement map $R_m \in \mathbb{R}^{H \times W}$ for each modality $m \in \{\text{RGB}, \text{3D}\}$, supervised by the pseudo mask $M^{*}$. In this manner, $f_i$ agrees with $f^*_i$ when normal and disagrees when anomalous, allowing the NSN to capture the feature discrepancy. The training objective combines a focal loss~\cite{focal} and a smooth $\ell_1$ loss ($\mathcal{L}_{reg}$):
% \vspace{-0.4em}
\begin{equation}
\mathcal{L}_{\text{NSN}} = \mathcal{L}_{\text{focal}} + \mathcal{L}_{reg}.
\end{equation}
The focal loss handles the binary classification between normal and anomalous regions, effectively addressing the class imbalance problem:
\begin{equation}
\mathcal{L}_{\text{focal}} = -\frac{1}{N}\sum_{i=1}^{N} \alpha_i (1-p_i)^\rho \log(p_i),
\end{equation}
where $p_i = \sigma(s_i)$ denotes the predicted anomaly probability at pixel $i$, obtained by applying a sigmoid function $\sigma(\cdot)$ to the refinement map $R_m$, and $\rho$ is the focusing factor. The smooth $L_1$ loss ensures regression consistency between the predicted refinement map $s_j = (R_m)_j$ and the pseudo mask $M^{*}_j$ at pixel $j$:
\begin{equation} 
\mathcal{L}_{reg} = \frac{1}{M}\sum_{j=1}^{M} 
\begin{cases} 
\frac{1}{2}(s_j - M_j^{*})^2 & \text{if } |s_j - M_j^{*}| < 1, \\ 
|s_j - M_j^{*}| - \frac{1}{2} & \text{otherwise}.
\end{cases}
\end{equation}

\begin{figure}[t]
\centering
\includegraphics[width=0.99\linewidth]{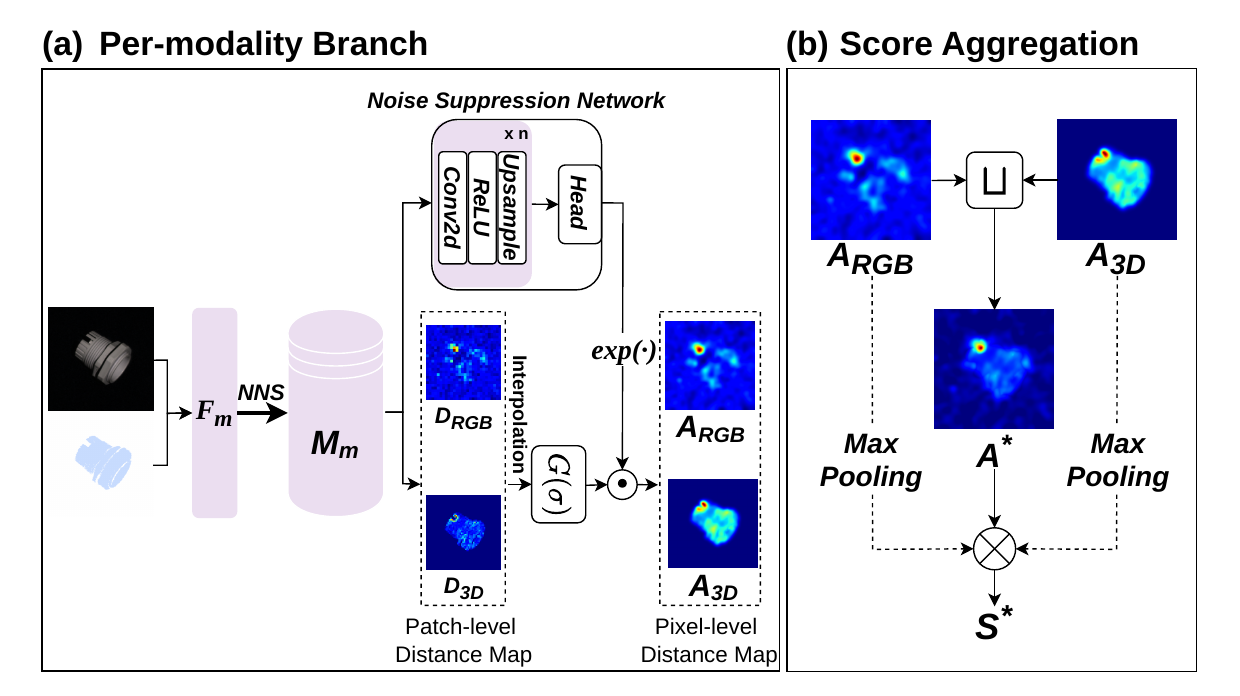}
\vspace{-1.0em}
\caption{
\textbf{Anomaly Prediction Stage.} \textbf{(a)} For each modality m ($m \in \{\text{RGB}, \text{3D}\}$), Nearest Neighbor Search (NNS) yields distance maps and matched features. These are processed through two parallel pathways. One upsamples distance maps with Gaussian smoothing ($G(\sigma)$) to produce coarse maps and the other leverages NSN to generate refinement maps. The final anomaly map combines both via Hadamard product. \textbf{(b)} The fusion operator ($\sqcup$) denotes z-score normalization followed by element-wise maximum fusion. Sample-level scores aggregate maximum values from all three anomaly maps.
}
\label{fig:inference}
\end{figure}

% \vspace{-0.5em}
\subsection{Anomaly Prediction}
\vspace{-0.1em}
As illustrated in Fig.~\ref{fig:inference}, our inference operates independently on each modality $m\in\{\text{RGB}, \text{3D}\}$
before fusing their outputs:

(i) We perform nearest neighbor search between input patch features and memory bank prototypes $\mathcal{M}_m$, yielding the $\ell_2$ distance:
\begin{equation}
d_i^m = \min_{f_j \in \mathcal{M}^m} \|f_i - f_j\|_2
\end{equation}
The distances are bilinearly upsampled with Gaussian smoothing to produce a coarse anomaly map $D_m$, following the standard PatchCore~\cite{patchcore} pipeline. 
Meanwhile, each patch feature $f_i$ and its matched prototype $f^*_i$ are concatenated and fed into NSN to produce a refinement map $R_m$.

(ii) The coarse map and refinement map are combined via Hadamard 
product~\cite{hadamard}, where $\exp(R_m)$ amplifies true anomaly signals and suppresses false anomaly signals:
\begin{equation}
\label{eq:modulate}
A_m = D_m \odot \exp(R_m), \quad R_m \in [-1,1]
\end{equation}

% (iii) After z-score normalization to align distance scales across modalities, the final fused map is obtained through element-wise maximum:
% \begin{equation}
% A^* = \max(\hat{A}_{\text{RGB}}, \hat{A}_{\text{3D}})
% \end{equation}

(iii) To align distance scales across modalities, we apply z-score normalization to each map $A_m$ using its own mean and standard deviation, yielding $\hat{A}_m$. The final fused map is then obtained through element-wise maximum:
\vspace{0.2em}
\begin{equation}
\label{eq:mapfusion}
A^* = \max(\hat{A}_{\text{RGB}}, \hat{A}_{\text{3D}})
\end{equation}

For sample-level detection, we leverage all three anomaly maps to capture modality-specific and complementary information. The maximum value from each of the three refined anomaly maps is extracted ($A_{\text{RGB}}$, $A_{\text{3D}}$, $A^*$) and aggregated as the final anomaly score:
\vspace{0.2em}
\begin{equation}
\label{eq:scoreagg}
S^* = \max(A_{\text{RGB}}) \times \max(A_{\text{3D}}) \times \max(A^*)
\end{equation}
\begin{table*}[t]
  \centering
  \footnotesize
  \caption{Multi-modal anomaly detection and segmentation results on (a) MVTec 3D-AD~\cite{mvtec3d}
  and (b) Eyecandies~\cite{eyecandies}. Best results in \textbf{bold}, second-best \underline{underlined}. ShapeGuided$^\dagger$ denotes baseline results from the original paper, reporting only the mean score.}
  \label{tab:main-results}

  %================= (a) MVTec 3D-AD =================
  \begin{subtable}{0.52\textwidth}
    \centering
    \caption{Results on MVTec 3D-AD}
    \label{tab:MVTec-Main}
    \setlength\tabcolsep{2.5pt}
    \renewcommand{\arraystretch}{0.8}
    \resizebox{\linewidth}{!}{%
    \begin{tabular}{c|l|*{10}{c}|c}
      \toprule
        \textbf{Metric} & \textbf{Method}
        & \textbf{Bag.} & \textbf{Cab.G.} & \textbf{Car.} & \textbf{Coo.} & \textbf{Dow.}
        & \textbf{Foa.} & \textbf{Pea.} & \textbf{Pot.} & \textbf{Rop.} & \textbf{Tir.}
        & \textbf{Mean} \\
      \midrule
      \multirow{7}{*}{\rotatebox{90}{\scriptsize \textbf{I-AUROC}}}
        & BTF~\cite{BTF}            & 91.8 & 74.8 & 96.7 & 88.3 & 93.2 & 58.2 & 89.6 & 91.2 & 92.1 & 88.6 & 86.5 \\
        & M3DM~\cite{M3DM}          & 99.4 & 90.9 & 97.2 & 97.6 & 96.0 & 94.2 & 97.3 & 89.9 & 97.2 & 85.0 & 94.5 \\
        & ShapeGuided~\cite{shapeguided} & 98.6 & 89.4 & 98.3 & \underline{99.1} & \underline{97.6} & 85.7 & \underline{99.0} & 96.5 & 96.0 & 86.9 & 94.7 \\
        & CFM~\cite{CFM}            & 98.8 & 87.5 & 98.4 & \textbf{99.2} & \textbf{99.7} & 92.4 & 96.4 & 94.9 & 97.9 & 95.0 & 96.0 \\
        & LSFA~\cite{LSFA}          & \textbf{100} & \underline{93.9} & 98.2 & 98.9 & 96.1 & 95.1 & 98.3 & 96.2 & \textbf{98.9} & \underline{95.1} & \underline{97.1} \\
        & G$^2$SF~\cite{g2sf}       & \underline{99.7} & 92.3 & \textbf{99.3} & 96.7 & 96.6 & \textbf{99.1} & \textbf{99.4} & \underline{98.8} & 96.6 & 92.2 & \underline{97.1} \\
        & Ours                      & \textbf{100} & \textbf{95.2} & \underline{99.1} & 98.5 & 96.7 & \underline{98.1} & 98.8 & \textbf{99.5} & \underline{98.7} & \textbf{99.5} & \textbf{98.4} \\
      \midrule
      \multirow{7}{*}{\rotatebox{90}{\scriptsize \textbf{AUPRO@30\%}}}
        & BTF~\cite{BTF}            & 97.6 & 96.7 & 97.9 & \underline{97.4} & \textbf{97.1} & 88.4 & 97.6 & \underline{98.1} & 95.9 & 97.1 & 96.4 \\
        & M3DM~\cite{M3DM}          & 97.0 & 97.1 & 97.9 & 95.0 & 94.1 & 93.2 & 97.7 & 97.1 & 97.1 & 97.5 & 96.4 \\
        & ShapeGuided~\cite{shapeguided} & 98.1 & 97.3 & \textbf{98.2} & 97.1 & 96.2 & \underline{97.8} & 98.1 & \textbf{98.3} & 97.4 & 97.5 & 97.6 \\
        & CFM~\cite{CFM}            & 98.0 & 96.6 & \textbf{98.2} & 94.7 & 95.9 & 96.7 & \underline{98.2} & \textbf{98.3} & \underline{97.6} & \underline{98.2} & 97.2 \\
        & LSFA~\cite{LSFA}          & \textbf{98.6} & 97.4 & \underline{98.1} & 94.6 & 92.5 & 94.1 & \textbf{98.3} & \textbf{98.3} & 97.4 & \textbf{98.3} & 96.8 \\
        & G$^2$SF~\cite{g2sf}       & \underline{98.2} & \underline{97.6} & \textbf{98.2} & \textbf{97.9} & \textbf{97.1} & 97.6 & \underline{98.2} & \textbf{98.3} & \textbf{97.8} & 98.1 & \textbf{97.9} \\
        & Ours                      & \underline{98.2} & \textbf{98.0} & \textbf{98.2} & 96.2 & \underline{96.3} & \textbf{98.1} & \textbf{98.3} & \textbf{98.3} & \textbf{97.8} & \textbf{98.3} & \underline{97.8} \\
      \midrule
      \multirow{6}{*}{\rotatebox{90}{\scriptsize \textbf{AUPRO@1\%}}}
        & BTF~\cite{BTF}            & 42.8 & 36.5 & 45.2 & 43.1 & 37.0 & 24.4 & 42.7 & 47.0 & 29.8 & 34.5 & 38.3 \\
        & M3DM~\cite{M3DM}          & 41.4 & 39.5 & 44.7 & 31.8 & \underline{42.2} & 33.5 & 44.4 & 35.1 & 41.6 & 39.8 & 39.4 \\
        & ShapeGuided$^\dagger$~\cite{shapeguided} & - & - & - & - & - & - & - & - & - & - & 45.6 \\
        & CFM~\cite{CFM}            & 45.9 & 43.1 & \underline{48.5} & \underline{46.9} & 39.4 & 41.3 & 46.8 & 48.7 & 46.4 & \underline{47.6} & 45.5 \\
        & G$^2$SF~\cite{g2sf}       & \underline{48.1} & \underline{44.3} & 48.4 & \textbf{47.1} & 41.0 & \underline{46.8} & \underline{48.7} & \underline{49.9} & \textbf{46.8} & 47.1 & \underline{46.8} \\
        & Ours                      & \textbf{53.7} & \textbf{48.0} & \textbf{49.1} & 45.9 & \textbf{42.8} & \textbf{48.9} & \textbf{50.4} & \textbf{50.7} & \underline{46.5} & \textbf{50.4} & \textbf{48.6} \\
      \bottomrule
    \end{tabular}}
  \end{subtable}%
  \hfill
  %================= (b) Eyecandies =================
  \begin{subtable}{0.47\textwidth}
    \centering
    \caption{Results on Eyecandies}
    \label{tab:eyecandies_results}
    \setlength\tabcolsep{0.8pt}
    \renewcommand{\arraystretch}{1.35}
    \resizebox{\linewidth}{!}{%
    \begin{tabular}{c|l|*{10}{c}|c}
      \toprule
        \textbf{Metric} & \textbf{Method}
        & \textbf{Can.C.} & \textbf{Cho.C.} & \textbf{Cho.P.} & \textbf{Con.} & \textbf{Gum.B.}
        & \textbf{Haz.T.} & \textbf{Lic.S.} & \textbf{Lol.} & \textbf{Mar.M.} & \textbf{Pep.C.}
        & \textbf{Mean} \\
      \midrule
      \multirow{5}{*}{\rotatebox{90}{\scriptsize \textbf{I-AUROC}}}
        & M3DM~\cite{M3DM}    & 62.4 & 95.8 & 95.8 & \textbf{100} & 88.6 & 75.8 & \textbf{94.9} & 83.6 & \textbf{100} & \textbf{100} & 89.7 \\
        & CFM~\cite{CFM}      & 68.0 & 93.1 & 95.2 & 88.0 & 86.5 & \textbf{78.2} & 91.7 & 84.0 & \underline{99.8} & \underline{96.2} & 88.1 \\
        & LSFA~\cite{LSFA}    & 67.0 & 95.4 & 96.1 & \textbf{100} & \textbf{91.3} & 76.7 & \underline{94.3} & 85.4 & \textbf{100} & \textbf{100} & \underline{90.6} \\
        & G$^2$SF~\cite{g2sf} & \underline{69.6} & \underline{96.3} & \underline{96.7} & \textbf{100} & \underline{90.7} & 70.1 & 93.9 & \underline{85.5} & 98.9 & \textbf{100} & 90.2 \\
        & Ours               & \textbf{73.1} & \textbf{97.1} & \textbf{98.4} & \underline{99.5} & 87.3 & \underline{77.3} & 90.7 & \textbf{88.6} & \underline{99.8} & \textbf{100} & \textbf{91.2} \\
      \midrule
      \multirow{4}{*}{\rotatebox{90}{\scriptsize \textbf{AUPRO@30\%}}}
        & M3DM~\cite{M3DM}    & 90.6 & \textbf{92.3} & 80.3 & \textbf{98.3} & 85.5 & 68.8 & \underline{88.0} & \underline{90.6} & \underline{96.6} & 95.5 & 88.2 \\
        & CFM~\cite{CFM}      & \underline{94.2} & 90.2 & \underline{83.1} & 96.5 & 87.5 & \textbf{76.2} & 79.1 & \textbf{91.3} & 93.9 & 94.9 & 88.7 \\
        & G$^2$SF~\cite{g2sf} & 92.8 & 89.7 & \textbf{84.3} & \underline{98.2} & \underline{89.0} & 68.7 & \textbf{88.8} & \textbf{91.3} & \textbf{97.1} & \underline{98.1} & \underline{89.8} \\
        & Ours               & \textbf{95.1} & \underline{92.0} & 81.9 & \underline{98.2} & \textbf{89.7} & \underline{76.1} & 85.6 & 90.2 & 93.7 & \textbf{98.3} & \textbf{90.1} \\
      \midrule
      \multirow{4}{*}{\rotatebox{90}{\scriptsize \textbf{AUPRO@1\%}}}
        & M3DM~\cite{M3DM}    & 16.6 & 38.8 & 32.9 & 48.6 & 31.5 & 13.1 & 32.3 & 25.8 & \underline{46.2} & 45.4 & 33.1 \\
        & CFM~\cite{CFM}      & \underline{22.9} & 30.7 & 34.5 & 38.9 & \underline{35.3} & \underline{18.8} & 33.3 & 23.6 & 45.5 & 42.8 & 33.5 \\
        & G$^2$SF~\cite{g2sf} & 17.1 & \textbf{41.6} & \textbf{37.7} & \underline{48.7} & 30.0 & 16.4 & \textbf{35.3} & \textbf{28.1} & \textbf{47.8} & \underline{47.9} & \underline{35.7} \\
        & Ours        & \textbf{30.5} & \underline{40.9} & \underline{37.6} & \textbf{51.3} & \textbf{38.7} & \textbf{20.4} & \underline{34.2} & \underline{27.5} & 42.6 & \textbf{52.5} & \textbf{37.6} \\
      \bottomrule
    \end{tabular}}
  \end{subtable}
\end{table*}

\vspace{-1em}
\section{Experiment}
\label{sec:Experiment}
\vspace{-0.5em}
\subsection{Experimental Settings}
\vspace{-0.5em}
% PD-REAL 부분 작성할 것.
\noindent{\textbf{Datasets.}}
We conduct experiments on MVTec 3D-AD~\cite{mvtec3d} and Eyecandies~\cite{eyecandies}.
MVTec 3D-AD contains 10 categories with 2,656 training, 294 validation, and 1,197 test samples, providing RGB images and organized point clouds. Eyecandies comprises 10 categories with 10,000 training, 1,000 validation, and 4,500 test samples, providing RGB images and depth maps under six lighting conditions. For Eyecandies, we obtain organized point clouds from the provided depth maps following prior work~\cite{M3DM} and use them as the 3D modality. To further assess generalizability, we additionally evaluate on PD-REAL~\cite{pdreal}. PD-REAL comprises 15 categories with six anomaly types, providing over 3,500 RGB images and point clouds captured by a RealSense camera under varying lighting conditions.

\noindent{\textbf{Evaluation Metrics.}}
As in prior works~\cite{BTF,M3DM,LSFA,g2sf,CFM,shapeguided}, we assess the image-level anomaly detection performance using the Area Under the Receiver Operator Curve (denoted as I-AUROC). For pixel-level anomaly segmentation, we employ pixel-level AUROC (P-AUROC) and the Area Under the Per-Region Overlap (AUPRO), with P-AUROC metrics reported in the supplementary material, Sec~C. For segmentation, we report AUPRO at False Positive Rate (FPR) thresholds of 30\% (denoted as AUPRO@30\%), and additionally evaluate at the more stringent 1\% FPR threshold (denoted as AUPRO@1\%) to demonstrate practical applicability in real-world industrial scenarios~\cite{CFM,g2sf,shapeguided}.

\noindent{\textbf{Implementation Detail.}}
We conduct all experiments on a single NVIDIA RTX A6000 48GB GPU. We use ViT-B/8 with DINO~\cite{dino} for RGB and PointMAE~\cite{pointmae} for 3D. Memory banks are constructed with 10\% prototype sampling~\cite{M3DM,g2sf,patchcore}. More details are in the supplementary material, Sec.~A.

\subsection{Main Results on Multi-Modal Benchmarks}
% \vspace{-0.2em}
We conduct experiments on MVTec 3D-AD~\cite{mvtec3d} and Eyecandies~\cite{eyecandies}, as reported in Tab.~\ref{tab:MVTec-Main} and Tab.~\ref{tab:eyecandies_results} respectively. On MVTec 3D-AD, our method outperforms previous leading approaches in sample-level detection, achieving a +1.3\% improvement in I-AUROC. For pixel-level segmentation, our method achieves a +1.8\% improvement in AUPRO@1\%, while remaining competitive at AUPRO@30\%. 
On Eyecandies, our method outperforms previous approaches on I-AUROC, AUPRO@30\%, and AUPRO@1\%, with notable gains in AUPRO@1\% (+1.9\%). 
These consistent improvements under stringent FPR constraints are particularly significant for industrial applications where minimizing false alarms is critical~\cite{unraveling,enhancingsvm}. As shown in Fig.~\ref{fig:qualitative_ablation}, the qualitative comparisons with previous works~\cite{g2sf,LSFA} demonstrate the effectiveness of our approach. Our method accurately localizes defect regions with sharp boundaries. We provide detailed analyses in Sec.~\ref{subsec:ablation}, and report per-modality scores (Sec.~C) and qualitative results (Sec.~D) in the supplementary material.

\begin{table}
\centering
\scriptsize
\setlength\tabcolsep{0.8pt}
\renewcommand{\arraystretch}{1.0}
\caption{Quantitative comparison under few-shot settings (5, 10, and 50 shots) on the MVTec 3D-AD dataset~\cite{mvtec3d}. The best results are highlighted in \textbf{bold} and the second-best results are \underline{underlined}. I-AUC denotes I-AUROC, PRO@30\% denotes AUPRO@30\%, and PRO@1\% denotes AUPRO@1\%.}
\label{tab:few_shot_results}
\begin{tabular}{l@{\hspace{3pt}}ccc@{\hspace{4pt}}ccc@{\hspace{4pt}}ccc}
\toprule
\multirow{2}{*}{\textbf{Method}} & \multicolumn{3}{c}{\textbf{5-shot}} & \multicolumn{3}{c}{\textbf{10-shot}} & \multicolumn{3}{c}{\textbf{50-shot}} \\ \cmidrule(lr){2-4} \cmidrule(lr){5-7} \cmidrule(lr){8-10}
 & {\tiny\textbf{I-AUC}} & {\tiny\textbf{PRO@30\%}} & {\tiny\textbf{PRO@1\%}} & {\tiny\textbf{I-AUC}} & {\tiny\textbf{PRO@30\%}} & {\tiny\textbf{PRO@1\%}} & {\tiny\textbf{I-AUC}} & {\tiny\textbf{PRO@30\%}} & {\tiny\textbf{PRO@1\%}} \\ \midrule
BTF~\cite{BTF} & 67.1 & 92.0 & 28.8 & 69.5 & 92.8 & 30.8 & 80.6 & 94.7 & 35.6 \\
M3DM~\cite{M3DM} & 82.2 & 93.7 & 33.0 & 84.5 & 94.3 & 35.5 & 90.7 & 95.5 & 38.7 \\
CFM~\cite{CFM} & 81.1 & \underline{94.9} & \underline{38.2} & 84.5 & \underline{95.4} & 39.8 & 90.6 & \underline{96.5} & \underline{43.1} \\
LSFA~\cite{LSFA} & 83.4 & 93.6 & -- & \underline{87.1} & 94.3 & -- & \underline{92.6} & 96.2 & -- \\
G$^2$SF~\cite{g2sf} & \underline{84.6} & 94.2 & 37.6 & 84.9 & 95.1 & \underline{40.0} & 91.4 & 95.9 & 43.0 \\ \midrule
\textbf{Ours} & \textbf{85.6} & \textbf{95.7} & \textbf{40.1} & \textbf{88.2} & \textbf{96.4} & \textbf{42.9} & \textbf{93.5} & \textbf{97.1} & \textbf{45.6} \\ \bottomrule
\end{tabular}
\end{table}

\vspace{-0.2em}
\subsection{Few-shot Results on MVTec 3D-AD}
\vspace{-0.2em}
Data scarcity is a critical constraint in industrial settings~\cite{CFM,scarcity1}. To assess robustness under this condition, we conduct few-shot experiments on MVTec 3D-AD~\cite{mvtec3d} with 5, 10, and 50 training samples per category. Following the previous work~\cite{CFM}, the $K$ samples of each category are taken as the first $K$ samples of the training set. In each $K$-shot setting, the entire pipeline---feature adaptation, memory bank construction, and NSN training---uses only these $K$ normal samples. All reported results are averaged over three training runs with different random seeds, with the $K$-shot subset held fixed. As shown in Tab.~\ref{tab:few_shot_results}, our method achieves state-of-the-art across all settings and metrics, demonstrating remarkable sample efficiency.
% Notably, on the segmentation metrics (AUPRO@30\% and AUPRO@1\%), our 5-shot model is competitive with the best 10-shot baselines, and our 10-shot model is competitive with the best 50-shot baselines, demonstrating remarkable sample efficiency.
\subsection{Comparison of Adaptation Strategies}
\label{subsec:adaptation_strategy}
\begin{table}
\centering
\scriptsize
\setlength{\tabcolsep}{4pt}
\renewcommand{\arraystretch}{1.0}
\caption{Quantitative comparison of feature adaptation strategies.}
\label{tab:adaptation}
\begin{tabular}{ccc}
\toprule
\textbf{Strategy} & \textbf{I-AUROC} & \textbf{AUPRO@30\%} \\
\midrule
Contrastive~\cite{LSFA} & \underline{97.1} & 96.8 \\
Reconstruction~\cite{MAE,pointmae} & 93.6 & \underline{97.3} \\
\midrule
Embedding Pred. & 95.1 & \textbf{97.7} \\
\textbf{+ VIC reg.} & \textbf{97.9} & \textbf{97.7} \\
\bottomrule
\end{tabular}
\end{table}
We compare our embedding prediction-based approach~\cite{ijepa,cjepa,pointjepa} against a contrastive learning-based approach~\cite{LSFA} and a reconstruction-based approach~\cite{MAE,pointmae}, to solely evaluate the impact of the feature adaptation strategy. All strategies share identical settings---same backbones~\cite{dino,pointmae}, feature extraction (single-scale for RGB, multi-scale for 3D), and memory bank-based procedure~\cite{patchcore} for anomaly detection and segmentation---differing only in the adaptation strategy.
As shown in Tab.~\ref{tab:adaptation}, both mask prediction-based strategies, reconstruction and embedding prediction, improve segmentation over contrastive adaptation (+0.5\% and +0.9\% in AUPRO@30\%, respectively). 
This aligns with the observation in~\cite{fcl} that spatial encoding benefits defect localization.
By predicting masked regions from surrounding context, mask prediction encourages learning local spatial dependencies among neighboring regions~\cite{enhancingjepa,maeanal}, which drives this improvement.
However, such locality does not equally benefit detection at the sample-level. 
% This gap reflects a difference in granularity. While pixel-level segmentation compares regions within a sample, where locality suffices to reveal how a region differs from its context, sample-level detection relies on a per-sample representative score (\eg, max-pooling) compared across samples, which locality alone does not guarantee.
Pixel-level segmentation compares regions within a sample, where locality suffices to reveal how a region differs from its context. Sample-level detection, in contrast, relies on a per-sample representative score (\eg, max-pooling) compared across samples, which locality alone does not guarantee.
We empirically find that VIC regularization, which prevents collapse and encourages each dimension to encode distinct information, improves separability between normal and anomalous samples.
This raises I-AUROC from 0.951 to 0.979 while maintaining segmentation, surpassing contrastive adaptation on both axes. We provide a detailed analysis of the impact of VIC regularization in the supplementary material (Sec.~E).

\subsection{Ablation Study}
\label{subsec:ablation}
% We conduct comprehensive ablation studies to validate our design choices and demonstrate the contribution of each component. All experiments are performed on MVTec 3D-AD dataset~\cite{mvtec3d}.
We conduct comprehensive ablation studies to analyze our design choices and demonstrate the contribution of each component. All experiments are performed on MVTec 3D-AD dataset~\cite{mvtec3d}.

\begin{table}
\centering
\scriptsize
\setlength{\tabcolsep}{2pt}
\renewcommand{\arraystretch}{1.1}
\caption{Comparison of batch-level and patch-level VIC regularization on MVTec 3D-AD.}
\label{tab:batch_patch}
\begin{tabular}{lcccc}
\toprule
\textbf{Granularity} & \textbf{I-AUROC} & \textbf{AUPRO@30\%} & \textbf{AUPRO@1\%} \\
\midrule
Batch-level & 97.6 & 97.3 & 46.5 \\
Patch-level & \textbf{98.1} & \textbf{97.7} & \textbf{47.6} \\
\bottomrule
\end{tabular}
\end{table}

\noindent\textbf{Granularity of VIC Regularization (Batch vs.\ Patch).} We validate our choice of applying VIC across patches within each sample rather than across samples within a batch.
This comparison isolates the adaptation stage alone, without the subsequent refinement stage (NSN).
The batch-level formulation~\cite{cjepa} applies mean pooling over the patches of each sample before VIC regularization, whereas our patch-level formulation applies it directly to each patch feature.
As shown in Tab.~\ref{tab:batch_patch}, we observe that VIC regularization is more effective when applied at the granularity at which anomalies are identified, \ie, per patch.
We provide a deeper analysis of the impact of VIC regularization in the supplementary material (Supp.~E.2).
% To understand this, we analyze the per-dimension variance and inter-dimensional covariance of the patch features for each modality.
% The patch-level formulation attains lower covariance in both modalities while keeping variance comparable.
% This stems from regularizing each patch feature directly rather than its mean-pooled representation.
% As a result, each patch feature carries more distinct, non-redundant information.
% Such patch-level distinctiveness benefits memory bank-based IAD, which judges anomalies by per-patch feature discrepancy.
\begin{table}[t]
    \centering
    \scriptsize
    \setlength{\tabcolsep}{4pt}
    \caption{Ablation study on feature adaptation (\textbf{Ada.}) and the noise suppression network (\textbf{NSN}).}
    \renewcommand{\arraystretch}{1.0}
    \begin{tabular}{ccccc}
    \toprule
    \textbf{Ada.} & \textbf{NSN} & \textbf{I-AUROC} & \textbf{AUPRO@30\%} & \textbf{AUPRO@1\%} \\
    \midrule
    \ding{55} & \ding{55} & 80.8 & 94.8 & 34.5 \\[3pt]
    \ding{51} & \ding{55} & 98.1 & 97.7 & 47.6 \\[3pt]
    % \ding{55} & \ding{51} & 0.791 & 0.828 & 0.205 \\[3pt]
    % \midrule
    \ding{51} & \ding{51} & \textbf{98.4} & \textbf{97.8} & \textbf{48.6} \\
    \bottomrule
    \end{tabular}
    \label{tab:ablation_components}
\end{table}

\begin{figure}[t]
    \centering
    \includegraphics[width=0.75\linewidth]{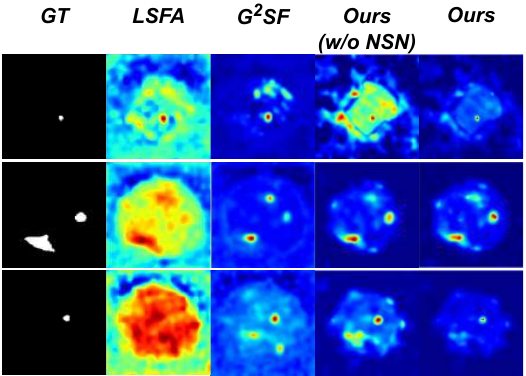}
    \caption{Qualitative results of feature adaptation and noise suppression network (\textbf{NSN}).}
    \label{fig:qualitative_ablation}
\end{figure}

\noindent\textbf{Impact of Feature Adaptation and Noise Suppression Network.} We analyze the contribution of each stage in our framework.
Tab.~\ref{tab:ablation_components} presents the ablation study on feature adaptation (\textbf{Ada.}) and the Noise Suppression Network (\textbf{NSN}).
Embedding prediction-based feature adaptation alone yields significant gains in both detection and segmentation, bridging the domain gap and enhancing normal–anomaly discriminability. 
% narrowing the claim of the contribution of NSN.
Integrating the refinement map of NSN improves the reliability of defect localization. In particular, it contributes a meaningful gain on AUPRO@1\% (+1.0\%), a stringent metric of practical importance in industrial settings.
% Qualitative로 옮기기 위한 연결고리 작업 필요.
As shown in Fig.~\ref{fig:qualitative_ablation}, interpolation-based anomaly maps ($\text{G}^2\text{SF}$~\cite{g2sf} and Ours w/o NSN) remain coarse and lack fine-grained details such as foreground shape. The refinement map from NSN addresses this by suppressing false positives and sharpening defect boundaries while preserving true anomaly \mbox{regions}.

\begin{table}[t]
\centering\footnotesize
\setlength{\tabcolsep}{4pt}
\renewcommand{\arraystretch}{0.9}
\caption{NSN refinement design comparison and its robustness on MVTec 3D-AD (multi-modal setting).}
\label{tab:nsn_des_rob}
\begin{subtable}{\linewidth}\centering
\caption{Refinement strategy}
\vspace{-0.3em}
\begin{tabular}{lccc}
\toprule
Strategy & I-AUROC & AUPRO@30\% & AUPRO@1\% \\
\midrule
Direct & 88.6 & 86.5 & 34.1 \\
\textbf{Modulate} & \textbf{98.4} & \textbf{97.8} & \textbf{48.6} \\
\bottomrule
\end{tabular}
\end{subtable}
\vspace{4pt}
\begin{subtable}{\linewidth}\centering
\caption{Synthesis robustness}
\begin{tabular}{lccc}
\toprule
Synthesis & I-AUROC & AUPRO@30\% & AUPRO@1\% \\
\midrule
Within-dataset & 98.5 & 97.7 & 48.2 \\
DTD & 98.4 & 97.8 & 48.6 \\
\bottomrule
\end{tabular}
\end{subtable}
\end{table}

% \begin{table}[t]
% \centering\footnotesize
% \setlength{\tabcolsep}{2pt}
% \renewcommand{\arraystretch}{0.9}
% \caption{Fusion strategy ablation on MVTec 3D-AD.
% PRO@30\% and PRO@1\% denote AUPRO at FPR thresholds of 30\% and 1\%, respectively.}
% \label{tab:fusion_comp}
% \begin{subtable}[t]{0.50\linewidth}\centering
% \caption{Map fusion (segmentation)}
% \begin{tabular}{lcc}
% \toprule
% Fusion & PRO@30\% & PRO@1\% \\
% \midrule
% Mean & 96.9 & 45.7 \\
% Hadamard & 97.4 & 48.3 \\
% \textbf{Max} & \textbf{97.8} & \textbf{48.6} \\
% \bottomrule
% \end{tabular}
% \end{subtable}
% \hfill
% \begin{subtable}[t]{0.47\linewidth}\centering
% \caption{Score aggregation (detection)}
% \begin{tabular}{lc}
% \toprule
% Aggregation & I-AUROC \\
% \midrule
% Sum & 87.3 \\
% Maximum & 97.5 \\
% \textbf{Product} & \textbf{98.4} \\
% \bottomrule
% \end{tabular}
% \end{subtable}
% \end{table}

\begin{table}[t]
\centering
\scriptsize
\setlength{\tabcolsep}{2pt}
\renewcommand{\arraystretch}{0.9}
\caption{Impact of memory bank size and layer selection (Single vs. Multi) on inference latency and anomaly detection performance.}
\label{tab:bank_size_comparison}
\resizebox{\columnwidth}{!}{%
\begin{tabular}{lccccc}
\toprule
\textbf{Method} & \textbf{Bank Size} & \textbf{I-AUROC} & \textbf{AUPRO@30\%} & \textbf{AUPRO@1\%} & \textbf{Latency} \\
\midrule
CFM~\cite{CFM} & \ding{55} & 96.0 & 97.2 & 45.5 & 0.28$\times$ \\
M3DM~\cite{M3DM} & 10\% & 94.5 & 96.4 & 39.4 & 2.85$\times$ \\
G$^2$SF~\cite{g2sf} & 10\% & 97.1 & \textbf{97.9} & 46.8 & 1.00$\times$ \\
\midrule
\multirow{3}{*}{\makecell{\textbf{Ours}\\\textbf{(Single)}}}
& 5\% & 98.0 & \underline{97.8} & 47.8 & \textbf{0.99$\times$} \\
& 10\% & 98.0 & 97.7 & 48.3 & 1.08$\times$ \\
& 20\% & 97.9 & \underline{97.8} & 48.0 & 1.22$\times$ \\
\midrule
\multirow{3}{*}{\makecell{\textbf{Ours}\\\textbf{(Multi)}}}
& 5\% & 98.1 & 97.6 & 48.3 & 1.16$\times$ \\
& 10\% & \textbf{98.4} & \underline{97.8} & \textbf{48.6} & 1.36$\times$ \\
& 20\% & \underline{98.3} & 97.7 & \underline{48.5} & 1.70$\times$ \\
\bottomrule
\end{tabular}%
}
\end{table}

\noindent\textbf{Refinement Design and Its Robustness.} Tab.~\ref{tab:nsn_des_rob} compares two strategies for refining the interpolation-based pixel-level anomaly map, which is coarse in nature. (a) Direct follows the refinement architecture of ReConPatch~\cite{reconpatch} to directly refine the coarse pixel-level map, whereas our Modulate uses the NSN refinement map to modulate it (Eq.~\ref{eq:modulate}). We observe that Direct even underperforms the coarse map without any refinement (Tab.~\ref{tab:ablation_components}). 
This is because directly refining a pixel-level map risks learning the appearance of synthetic anomalies, which differs explicitly from real defects. It thus fails to generalize to real anomalies at inference. In contrast, Modulate preserves the coarse distance-based map and selectively amplifies true anomaly signals, yielding sharper maps (Fig.~\ref{fig:qualitative_ablation}). (b) For RGB, pseudo-anomalies are synthesized using external DTD textures. 
We replace them with within-dataset synthesis, as in 3D. 
This yields nearly identical results across all metrics. 
Together, (a) and (b) indicate that NSN learns the feature discrepancy between each query patch feature and its nearest normal prototype, rather than the trivial appearance of synthetic anomalies, making it generalizable and robust.

% \noindent\textbf{Fusion Strategy.} We analyze our late-fusion strategy for combining per-modality outputs, separately for segmentation and detection (Tab.~\ref{tab:fusion_comp}).
% (a) For pixel-level map fusion, we compare mean, Hadamard product~\cite{hadamard}, and element-wise maximum after z-score normalization. We observe that maximum fusion performs best (Tab.~\ref{tab:fusion_comp}~(a)), preserving an anomalous response whenever either modality detects it.
% (b) For sample-level score aggregation, we compare sum, maximum, and product. Product performs best, surpassing maximum by +0.9\% in I-AUROC (Tab.~\ref{tab:fusion_comp}~(b)). 
% While maximum, acting as an OR-gate, remains competitive (0.975 vs.\ 0.971 for G$^2$SF~\cite{g2sf}), product acts as an AND-gate that requires cross-modal agreement, suppressing single-modality false positives. Such AND-gate behavior is more beneficial for detection than propagating every modality-specific activation.

% \vspace{-0.45em}
\noindent\textbf{Memory Bank Size and Latency Analysis.}
Since industrial deployment depends not only on accuracy but also on inference latency, we analyze the trade-off between performance and latency across varying memory bank sizes, as shown in Tab.~\ref{tab:bank_size_comparison}.
We use G$^2$SF~\cite{g2sf}, which meets practical industrial requirements, as our latency baseline (1.00$\times$) and report relative latency measured under identical conditions.
For the multi-scale layer selection configuration, 10\% bank size achieves the best performance but incurs 1.36$\times$ latency compared to G$^2$SF, primarily due to processing high-dimensional features. We further evaluate the single-scale configuration (using only the final layer for both RGB and 3D encoders) across varying memory bank sizes.
With 5\% memory bank, the single-scale configuration outperforms G$^2$SF by +0.9\% on I-AUROC and +1.0\% on AUPRO@1\% while maintaining competitive inference time (0.99$\times$).
Despite the trade-off between accuracy and latency, our single-scale configuration (5\%) attains competitive latency while surpassing the state-of-the-art baseline~\cite{g2sf} in I-AUROC and AUPRO@1\%.

\begin{table}[t]
\centering
\scriptsize
\caption{Results on PD-REAL~\cite{pdreal} in the multi-modal setting, averaged over
all 15 categories. Best results in bold, second-best underlined. \textbf{Ada.} denotes our feature adaptation stage and \textbf{NSN} the noise suppression network.}
\label{tab:pdreal}
\begin{tabular}{lccc}
\toprule
Method & I-AUROC & AUPRO@30\% & AUPRO@1\% \\
\midrule
G$^2$SF~\cite{g2sf}   & 94.9 & 94.7 & 43.1 \\
\midrule
Ours (Ada. only)      & \textbf{96.2} & \underline{97.5} & \underline{46.7} \\
Ours (Ada. + NSN)     & \underline{95.6} & \textbf{97.7} & \textbf{48.5} \\
\bottomrule
\end{tabular}
\end{table}

\subsection{Generalization to an Additional Benchmark}
\label{sec:pdreal}
% To further assess generalizability, we compare against G$^2$SF~\cite{g2sf}, a recent state-of-the-art work, on PD-REAL~\cite{pdreal}. 
We compare against G$^2$SF~\cite{g2sf}, a recent state-of-the-art work, on PD-REAL~\cite{pdreal}.
As shown in Tab.~\ref{tab:pdreal}, EPAR outperforms G$^2$SF across all metrics, with the largest margin on AUPRO@1\% (+5.4\%). 
The contribution of NSN is also evident here, yielding a meaningful gain in AUPRO@1\% that supports the improved reliability of defect localization.
Experimental settings on PD-REAL for both G$^2$SF and EPAR are provided in the supplementary material (Sec.~A).
\vspace{-1em}
% \vspace{-0.9em}
\vspace{0.55em}
\section{Conclusion}
\vspace{-0.3em}
In this work, we propose EPAR, a comprehensive multi-modal industrial anomaly detection framework. 
Our approach demonstrates that embedding prediction combined with patch-level VIC regularization effectively adapts pretrained encoders for industrial anomaly detection.
Building upon the adapted features, a Noise Suppression Network trained on pseudo-anomalies further refines coarse anomaly maps into fine-grained ones with sufficient detail, leading to reliable defect localization.
EPAR demonstrates strong performance across MVTec 3D-AD, Eyecandies, and PD-REAL, and remains effective under few-shot settings.
However, our analysis reveals a trade-off between accuracy and efficiency: multi-scale feature extraction achieves optimal performance but incurs additional latency, while single-scale extraction maintains competitive inference speed with slightly reduced accuracy. In addition, the benefit of our refinement stage is concentrated on pixel-level defect localization. We hope this work inspires further research on balancing accuracy and efficiency as well as on refinement strategies that further strengthen sample-level detection, potentially through cross-modal interaction during adaptation and refinement.
\section*{Acknowledgment}
This work was supported by Institute of Information \& Communications Technology Planning \& Evaluation (IITP) grant funded by the Korea government (MSIT) (RS-2026-25523906, Hyper-scale Industrial AI Research Support (R\&D) Program, Development of an industry-specified multimodal hyperscale foundation model, No. RS-2022-II220608/2022-0-00608, Artificial Intelligence research about multimodal interactions for empathetic conversations with humans, No. RS-2025-25442824, AI Star Fellowship Program (Ulsan National Institute of Science and Technology), \& No. RS-2020-II201336, Artificial Intelligence graduate school support (UNIST)).
% \newpage
{
    \small
    \bibliographystyle{ieeenat_fullname}
    \bibliography{main}
}

%%%%%%%%%%%%%%%%%%%%%%%%%%%%%%%%%%%%%%%%%%%%%%%%%%%%%%%%%%%%%%%%%%%%%%%%%%%%%%
%%%%%%%%%%%%%%%%%%%%%%%%%%%  SUPPLEMENTARY MATERIAL  %%%%%%%%%%%%%%%%%%%%%%%%%%
%%%%%%%%%%%%%%%%%%%%%%%%%%%%%%%%%%%%%%%%%%%%%%%%%%%%%%%%%%%%%%%%%%%%%%%%%%%%%%
% Start the supplementary material on a fresh page (flushes all pending floats).
\clearpage

% Supplementary title block (spans both columns; no page number on this page)
\twocolumn[{%
  \null\vskip .375in
  \begin{center}
    {\Large \bf Supplementary Material: Embedding Predictive Adaptation and Refinement for Multi-Modal Industrial Anomaly Detection \par}
    \vspace*{24pt}
  \end{center}
}]
% \thispagestyle{empty}

% --- Numbering for the supplementary material: Sec. A, Table A, Figure A, ...
% \theH* keep hyperref anchors unique w.r.t. the main paper.
\setcounter{section}{0}
\renewcommand{\thesection}{\Alph{section}}
\renewcommand{\thesubsection}{\Alph{section}.\arabic{subsection}}
\renewcommand{\theHsection}{supp.\Alph{section}}
\renewcommand{\theHsubsection}{supp.\Alph{section}.\arabic{subsection}}
\setcounter{table}{0}
\setcounter{figure}{0}
\renewcommand{\thetable}{\Alph{table}}
\renewcommand{\thefigure}{\Alph{figure}}
\renewcommand{\theHtable}{supp.\Alph{table}}
\renewcommand{\theHfigure}{supp.\Alph{figure}}
\setcounter{equation}{0}
\renewcommand{\theequation}{S\arabic{equation}}
\renewcommand{\theHequation}{supp.\arabic{equation}}
\setcounter{algorithm}{0}
\providecommand{\theHalgorithm}{}
\renewcommand{\theHalgorithm}{supp.\arabic{algorithm}}

% --- Caption / float spacing used in the supplementary material
\captionsetup{skip=6pt}                                  % gap between float and its caption
\setlength{\textfloatsep}{12pt plus 2pt minus 2pt}       % single-column floats <-> text
\setlength{\dbltextfloatsep}{14pt plus 2pt minus 2pt}    % full-width (table*) floats <-> text
\setlength{\intextsep}{10pt plus 2pt minus 2pt}          % in-place [H] floats <-> text
\setlength{\floatsep}{10pt plus 2pt minus 2pt}           % between stacked floats

\section*{Overview of the supplementary material}

This supplementary material provides additional details and experimental results to complement the main paper. The contents are organized as follows:

\begin{itemize}
    \item[$\bullet$] \textbf{Sec.~\ref{supp:sec:implement_detail}}: Detailed implementation details including target block sampling strategy for both RGB and 3D modalities, and experimental settings for feature adaptation and noise suppression network training.
    \item[$\bullet$] \textbf{Sec.~\ref{supp:sec:training_algo}}: Pseudo-code for the training procedures of each stage: feature adaptation stage and noise suppression network training stage.
    \item[$\bullet$] \textbf{Sec.~\ref{supp:sec:quantitative_results}}: Additional experiments including per-category analysis, P-AUROC results, generalization to uni-modal benchmarks, a detailed analysis of the refinement process across modalities, a sensitivity analysis on the VIC regularization coefficient, and a memory overhead analysis.
    \item[$\bullet$] \textbf{Sec.~\ref{supp:sec:qualitative_res}}: Qualitative visualization of anomaly maps across diverse categories on MVTec 3D-AD and Eyecandies datasets.
    \item[$\bullet$] \textbf{Sec.~\ref{supp:sec:feature_analysis}}: In-depth analysis of spatial relationship encoding and VIC regularization effects.
\end{itemize}

%%%%%%%%%%%%%%%%%%%%%%%%%%%%%%%%%%%%%%%%%%%%%%%%%%%%%%%%%%%%%
% SECTION A: Implementation Details
%%%%%%%%%%%%%%%%%%%%%%%%%%%%%%%%%%%%%%%%%%%%%%%%%%%%%%%%%%%%%
\section{Implementation Details}
\label{supp:sec:implement_detail}
\paratitle{Target Block Sampling Strategy.} We provide detailed descriptions of the target block sampling procedures for both RGB and 3D modalities. Following I-JEPA~\cite{ijepa} for RGB and Point-JEPA~\cite{pointjepa} for 3D point clouds, our sampling strategy selects multiple spatially-contiguous target blocks from semantically meaningful foreground regions. We first obtain the foreground region from the 3D point cloud, where we estimate the background plane with RANSAC~\cite{RANSAC} following~\cite{M3DM,CFM} and remove any point within 0.005 distance. Since each RGB pixel corresponds to a 3D point at the same location, we apply the same foreground region to the RGB image. For RGB, we then further refine the foreground patches within this region. We convert the RGB image to grayscale and patchify it, and classify a patch as background if the proportion of dark pixels (intensity below a threshold, e.g., 30) exceeds a ratio (e.g., 0.9), retaining the remaining patches as foreground. For 3D, since the background plane is already removed, the point groups formed by Farthest Point Sampling and KNN grouping, following~\cite{pointnet2}, naturally lie on the foreground, so no further foreground refinement is needed. From the identified foreground regions, we sample $M=4$ spatially-contiguous target blocks, each covering 15--20\% of the foreground patches. For 3D point clouds, following Point-JEPA~\cite{pointjepa}, we additionally sample 40--75\% of the remaining patches as the context input. The context input for each modality is obtained by excluding the target block regions.

\paratitle{Experimental Settings for Feature Adaptation.} 
We use AdamW~\cite{adamw} as the optimizer, with weight decay increased from 0.01 to 0.1 following a cosine schedule. The learning rate follows a warmup-cosine annealing schedule, linearly increasing during the first warmup epoch and then decaying to the base learning rate.
For the EMA update, the momentum starts at 0.995 and linearly increases to 1.0 over training. 
We adapt the encoders per category, tuning the hyperparameters for each category on a validation set using synthesized pseudo-anomalies.

\paratitle{Experimental Settings for Noise Suppression Network.} For NSN training, we use AdamW optimizer~\cite{adamw} as in the feature adaptation stage. The training is conducted for 5 epochs with a fixed learning rate of $10^{-5}$ or $10^{-4}$. The loss function combines Focal Loss~\cite{focal} and Smooth $\ell_1$ Loss.

\begin{algorithm*}[t]
\caption{Feature Adaptation Stage}
\label{supp:alg:adapt}
\begin{algorithmic}[1]
\Require $\mathcal{D} = \{(\mathbf{I}_i, \mathbf{P}_i)\}_{i=1}^{N}$,\;
         $E$: the number of epochs,\; $M$: the number of target blocks,\;
         context encoder $f^m$,\; target encoder $\bar{f}^m \leftarrow \texttt{deepcopy}(f^m)$,\;
         predictor $\Phi^m$,\; learnable mask tokens $\{\texttt{Token}^m_k\}_{k=1}^{M}$,\;
         {\small $m \in \{\text{RGB},\,\text{3D}\}$}
\For{$i = 1, \dots, E$}
  \For{$j = 1, \dots, N$}
    \State $\mathbf{C}^m \leftarrow f^m(c^m_j)$
           \hfill\Comment{$c^m_j$: context input of $j$-th data}
    \State $\mathbf{T}^m \leftarrow \texttt{stop\_grad}(\bar{f}^m(t^m_j))$
           \hfill\Comment{$t^m_j$: target input of $j$-th data}
    \State $\hat{\mathbf{T}}^m \leftarrow$ regions of $\mathbf{T}^m$ corresponding to $c^m_j$
    \State $\{\mathcal{T}^m_k\}_{k=1}^{M} \leftarrow$ a set of patch positions of $M$ sampled target blocks
    \State $\mathcal{L}^m_{\text{JEPA}} \leftarrow
           \dfrac{1}{M}\displaystyle\sum_{k=1}^{M}
           \texttt{SmoothL1}\!\bigl(
             \Phi^m([\mathbf{C}^m;\,\texttt{Token}^m_k]),\;
             \mathbf{T}^m[\mathcal{T}^m_k]
           \bigr)$
    \State $\mathcal{L}^m_{\text{VIC}} \leftarrow
           15\,\mathcal{L}_{\text{invar}}(\mathbf{C}^m,\hat{\mathbf{T}}^m) +
           15\,\mathcal{L}_{\text{var}}(\mathbf{C}^m) +
           \mathcal{L}_{\text{cov}}(\mathbf{C}^m)$
    \State $f^m,\, \Phi^m \overset{\text{optim}}{\leftarrow}
           \mathcal{L}^m_{\text{JEPA}} + \lambda\,\mathcal{L}^m_{\text{VIC}}$
    \State $\bar{f}^m \leftarrow \mu\,\bar{f}^m + (1-\mu)\,f^m$
           \hfill\Comment{EMA update}
  \EndFor
\EndFor
\end{algorithmic}
\end{algorithm*}

\begin{algorithm*}[t]
\caption{Noise Suppression Network Training Stage}
\label{supp:alg:nsn}
\begin{algorithmic}[1]
\Require Normal set $\mathcal{D}^{+}_m = \{x^m_i\}_{i=1}^{N}$,\;
         pseudo-anomaly set $\mathcal{D}^{-}_m = \{(\tilde{x}^m_i,\, \mathbf{m}^*_i)\}_{i=1}^{N}$,\;
         frozen adapted encoder $f^m_{\text{adapt}}$,\; NSN $\mathcal{N}^m$,\; epochs $E$,\;
         {\small $m \in \{\text{RGB},\,\text{3D}\}$}
\Statex \hspace*{-\algorithmicindent}\textbf{Memory Bank Construction}
\State $\mathcal{M}^m \overset{\text{coreset}}{\longleftarrow}
       \bigl\{\,f^m_{\text{adapt}}(x^m_i)\,\bigr\}_{i=1}^{N}$
       \hfill\Comment{prototype sampling from $\mathcal{D}^{+}_m$}
\Statex \hspace*{-\algorithmicindent}\textbf{NSN Training}
\For{$i = 1, \dots, E$}
  \For{$j = 1, \dots, N$}
    \State $f_j \leftarrow f^m_{\text{adapt}}(\tilde{x}^m_j)$
           \hfill\Comment{pseudo-anomaly feature}
    \State $f^*_j \leftarrow \arg\min_{f_k \in \mathcal{M}^m} \|f_j - f_k\|_2$
           \hfill\Comment{nearest normal prototype}
    \State $R^m_j \leftarrow \mathcal{N}^m([f_j;\, f^*_j])$
           \hfill\Comment{predicted refinement map}
    \State $\mathcal{L}^m_{\text{NSN}} \leftarrow
           \mathcal{L}_{\text{focal}}(R^m_j,\, \mathbf{m}^*_j) +
           \mathcal{L}_{\text{reg}}(R^m_j,\, \mathbf{m}^*_j)$
    \State $\mathcal{N}^m \overset{\text{optim}}{\leftarrow} \mathcal{L}^m_{\text{NSN}}$
  \EndFor
\EndFor
\end{algorithmic}
\end{algorithm*}

\paratitle{Experimental Settings on PD-REAL dataset.}
PD-REAL~\cite{pdreal} provides RGB images and point clouds captured with a RealSense camera under varying lighting across 15 categories with six anomaly types. We follow the official train/test split for feature adaptation, NSN training, and evaluation. None of the baselines report results on PD-REAL. We therefore run both ours (EPAR) and G$^2$SF~\cite{g2sf}. For EPAR, we use the same architecture and objective as in our experiments on MVTec 3D-AD and Eyecandies --- ViT-B/8 with DINO~\cite{dino} for RGB and PointMAE~\cite{pointmae} for 3D, adapting the intermediate layers $k \in \{4, 8, 12\}$ of both encoders, and constructing the memory bank with 10\% prototype sampling. Feature adaptation and NSN training are performed per category. For G$^2$SF~\cite{g2sf}, we use the authors' official implementation with the configuration released in their repository.

%%%%%%%%%%%%%%%%%%%%%%%%%%%%%%%%%%%%%%%%%%%%%%%%%%%%%%%%%%%%%
% SECTION B: Training Algorithms
%%%%%%%%%%%%%%%%%%%%%%%%%%%%%%%%%%%%%%%%%%%%%%%%%%%%%%%%%%%%%
\section{Training Algorithms}
\label{supp:sec:training_algo}
We describe the training procedure of each stage via pseudo-code. As the procedure is identical across modalities, we present a unified description applicable to both $m \in \{\text{RGB}, \text{3D}\}$.

Algorithm~\ref{supp:alg:adapt} summarizes the Feature Adaptation Stage. For each training sample, the context encoder $f^m$ processes the masked context input $c^m_j$, while the EMA-updated target encoder $\bar{f}^m$ provides stable target representations $\mathbf{T}^m$ from the full input $t^m_j$. The predictor $\Phi^m$ takes the concatenation of context features and learnable mask tokens to predict $M$ target blocks, supervised by the smooth $\ell_1$ loss $\mathcal{L}_{\text{JEPA}}$. VIC regularization $\mathcal{L}_{\text{VIC}}$ is jointly applied at the patch-level context features. All processes are applied independently at each intermediate layer $k \in \{4, 8, 12\}$, enabling multi-scale feature adaptation.

\begin{table*}[t]
\centering
\caption{I-AUROC for anomaly detection of all categories on MVTec 3D-AD.}
\label{supp:tab:supp_iauroc}
\resizebox{\linewidth}{!}{%
\scriptsize
\begin{tabular}{@{}c|lc|cccccccccc|c@{}}
\toprule
Modality & Method & yr & Bagel & Cable Gland & Carrot & Cookie & Dowel & Foam & Peach & Potato & Rope & Tire & Mean \\
\midrule
\multirow{5}{*}{\rotatebox{90}{RGB}}
& M3DM~\cite{M3DM} & 23 & 94.4 & 91.8 & 89.6 & 74.9 & 95.9 & 76.7 & 91.9 & \underline{64.8} & \underline{93.8} & 76.7 & 85.0 \\
& ShapeGuided~\cite{shapeguided} & 23 & 91.1 & \underline{93.6} & 88.3 & 66.2 & \textbf{97.4} & \underline{77.2} & 78.5 & 64.1 & 88.4 & 70.6 & 81.5 \\
& LSFA~\cite{LSFA} & 24 & \underline{95.1} & 92.0 & \underline{91.1} & \underline{76.2} & \underline{96.1} & 77.0 & \underline{93.0} & \textbf{67.5} & \underline{93.8} & \textbf{78.7} & \underline{86.1} \\
& G$^2$SF~\cite{g2sf} & 25 & 94.5 & \textbf{93.9} & \textbf{91.4} & 73.3 & 93.9 & 76.8 & \textbf{94.4} & 62.1 & 93.0 & 75.5 & 84.9 \\
& Ours & - & \textbf{97.5} & \underline{93.6} & 82.7 & \textbf{85.9} & 94.1 & \textbf{82.1} & 88.6 & 64.6 & \textbf{96.6} & \underline{75.8} & \textbf{86.2} \\
\midrule
\multirow{6}{*}{\rotatebox{90}{3D}}
& BTF~\cite{BTF} & 23 & 82.5 & 55.1 & 95.2 & 79.7 & 88.3 & 58.2 & 75.8 & 88.9 & 92.9 & 65.3 & 78.2 \\
& M3DM~\cite{M3DM} & 23 & 94.1 & 65.1 & 96.5 & 96.9 & 90.5 & 76.0 & 88.0 & \underline{97.4} & 92.6 & 76.5 & 87.4 \\
& ShapeGuided~\cite{shapeguided} & 23 & \underline{98.3} & \underline{68.2} & 97.8 & \textbf{99.8} & \textbf{96.0} & 73.7 & \textbf{99.3} & 97.9 & 96.6 & 87.1 & 91.6 \\
& LSFA~\cite{LSFA} & 24 & \textbf{98.6} & 66.9 & 97.3 & \underline{99.0} & \underline{95.0} & \underline{80.2} & 96.1 & 96.4 & \underline{96.7} & \underline{94.4} & \underline{92.1} \\
& G$^2$SF~\cite{g2sf} & 25 & 95.1 & 63.7 & \underline{98.1} & 95.2 & 83.7 & 78.6 & 92.3 & 93.0 & 85.8 & 68.8 & 85.4 \\
& Ours & - & \textbf{98.6} & \textbf{82.9} & \textbf{98.9} & 99.2 & \underline{95.0} & \textbf{83.7} & \underline{99.2} & 97.1 & \textbf{98.6} & \textbf{99.1} & \textbf{95.2} \\
\bottomrule
\end{tabular}%
}
\end{table*}

\begin{table*}[t]
\centering
\caption{AUPRO@30\% for anomaly segmentation of all categories on MVTec 3D-AD.}
\label{supp:tab:supp_aupro30}
\resizebox{\linewidth}{!}{%
\scriptsize
\begin{tabular}{@{}c|lc|cccccccccc|c@{}}
\toprule
Modality & Method & yr & Bagel & Cable Gland & Carrot & Cookie & Dowel & Foam & Peach & Potato & Rope & Tire & Mean \\
\midrule
\multirow{5}{*}{\rotatebox{90}{RGB}}
& M3DM~\cite{M3DM} & 23 & 95.2 & \underline{97.2} & \underline{97.3} & 89.1 & 93.2 & 84.3 & 97.0 & 95.6 & 96.8 & 96.6 & 94.2 \\
& ShapeGuided~\cite{shapeguided} & 23 & 94.6 & \underline{97.2} & 96.0 & \underline{91.4} & \underline{95.8} & 77.6 & 93.7 & 94.9 & 95.6 & 95.7 & 93.3 \\
& LSFA~\cite{LSFA} & 24 & \underline{95.7} & \textbf{97.6} & 97.0 & 91.2 & 93.4 & \underline{85.1} & 96.0 & \underline{95.7} & \underline{97.0} & 96.1 & 94.5 \\
& G$^2$SF~\cite{g2sf} & 25 & 95.4 & 97.3 & \textbf{97.4} & 88.7 & \textbf{97.4} & 84.7 & \underline{97.1} & \underline{95.7} & 96.5 & \underline{97.0} & \underline{94.7} \\
& Ours & - & \textbf{96.6} & 96.0 & 96.9 & \textbf{94.5} & 95.0 & \textbf{86.8} & \textbf{97.3} & \textbf{96.2} & \textbf{97.4} & \textbf{96.9} & \textbf{95.4} \\
\midrule
\multirow{6}{*}{\rotatebox{90}{3D}}
& BTF~\cite{BTF} & 23 & 97.3 & 87.9 & \textbf{98.2} & 90.6 & 89.2 & 73.5 & 97.7 & 98.2 & 95.6 & 96.1 & 92.4 \\
& M3DM~\cite{M3DM} & 23 & 94.3 & 81.8 & 97.7 & 88.2 & 88.1 & 74.3 & 95.8 & 97.4 & 95.0 & 92.9 & 90.6 \\
& ShapeGuided~\cite{shapeguided} & 23 & \textbf{97.4} & 87.1 & \underline{98.1} & \underline{92.4} & 89.8 & \underline{77.3} & 97.8 & \underline{98.3} & 95.5 & 96.9 & 93.1 \\
& LSFA~\cite{LSFA} & 24 & \textbf{97.4} & \underline{88.7} & \underline{98.1} & 92.1 & \underline{90.1} & \underline{77.3} & \underline{98.2} & \underline{98.3} & \underline{95.9} & \underline{98.1} & \underline{93.4} \\
& G$^2$SF~\cite{g2sf} & 25 & 96.0 & 80.6 & 97.7 & 89.9 & 92.9 & 76.1 & 97.1 & 97.6 & 94.6 & 92.7 & 91.5 \\
& Ours & - & \underline{97.3} & \textbf{94.0} & \underline{98.1} & \textbf{94.9} & 83.0 & \textbf{84.4} & \textbf{98.3} & \textbf{98.4} & \textbf{95.9} & \textbf{98.2} & \textbf{94.3} \\
\bottomrule
\end{tabular}%
}
\end{table*}

Algorithm~\ref{supp:alg:nsn} summarizes the Noise Suppression Network Training Stage. We first construct a memory bank $\mathcal{M}^m$ by storing patch-level normal features from $\mathcal{D}^{+}_m$ via prototype sampling following PatchCore~\cite{patchcore}, where features from intermediate layers $k \in \{4, 8, 12\}$ are concatenated to form multi-scale representations. During training, each pseudo-anomaly feature $f_j$ is paired with its nearest normal prototype $f^*_j$ retrieved from $\mathcal{M}^m$ via Nearest Neighbor Search (NNS), and the concatenated pair $[f_j;\, f^*_j]$ is passed into NSN $\mathcal{N}^m$ to produce a pixel-level refinement map $R^m_j$. The network is jointly supervised by focal loss $\mathcal{L}_{\text{focal}}$ for binary classification of normal and anomalous regions, and smooth $\ell_1$ loss $\mathcal{L}_{\text{reg}}$ for regression consistency against the pseudo mask $\mathbf{m}^*_j$.

%%%%%%%%%%%%%%%%%%%%%%%%%%%%%%%%%%%%%%%%%%%%%%%%%%%%%%%%%%%%%
% SECTION C: Quantitative Results
%%%%%%%%%%%%%%%%%%%%%%%%%%%%%%%%%%%%%%%%%%%%%%%%%%%%%%%%%%%%%
\section{Quantitative Results}
\label{supp:sec:quantitative_results}
In this section, we provide comprehensive quantitative analyses to supplement the main paper. We show per-category breakdowns, additional metrics, and further experiments on our design choices.

\paratitle{Per-Category Analysis.} Tab.~\ref{supp:tab:supp_iauroc} and Tab.~\ref{supp:tab:supp_aupro30} present per-category I-AUROC and AUPRO@30\% results for RGB and 3D modalities, respectively.
For RGB, our method achieves 86.2 I-AUROC and 95.4 AUPRO@30\% on average across all categories, outperforming prior methods~\cite{M3DM,shapeguided,LSFA,g2sf}.
For 3D, we observe more substantial gains, achieving 95.2 I-AUROC and 94.3 AUPRO@30\% averaged across all categories (+3.1\% and +0.9\% over LSFA~\cite{LSFA}, respectively).
These results demonstrate that our approach is particularly effective for 3D modality, showing significant improvements over competitive baselines.

\begin{table}[t]
\begin{minipage}[t]{0.48\textwidth}
\centering
\caption{P-AUROC for anomaly segmentation on MVTec 3D-AD. Results are reported for each modality.}
\label{supp:tab:pauroc_modality}
\scriptsize
\setlength{\tabcolsep}{4pt}
\begin{tabular}{@{}l|ccc@{}}
\toprule
Method & RGB & 3D & Multi-Modal \\
\midrule
BTF~\cite{BTF} & - & \underline{97.8} & 99.2 \\
M3DM~\cite{M3DM} & \underline{98.7} & 97.0 & 99.2 \\
ShapeGuided~\cite{shapeguided} & - & - & \underline{99.6} \\
CFM~\cite{CFM} & - & - & 99.3 \\
LSFA~\cite{LSFA} & \underline{98.7} & \textbf{98.1} & 99.3 \\
G$^2$SF~\cite{g2sf} & \textbf{98.8} & 97.5 & \textbf{99.7} \\
\midrule
Ours & \textbf{98.8} & 97.5 & \underline{99.6} \\
\bottomrule
\end{tabular}
\end{minipage}%
\hfill
\begin{minipage}[t]{0.48\textwidth}
\centering
\caption{P-AUROC under few-shot settings on MVTec 3D-AD. Multi-modal results are reported.}
\label{supp:tab:pauroc_fewshot}
\scriptsize
\setlength{\tabcolsep}{4pt}
\begin{tabular}{@{}l|ccc@{}}
\toprule
Method & 5-shot & 10-shot & 50-shot \\
\midrule
BTF~\cite{BTF} & 98.0 & 98.3 & 98.9 \\
M3DM~\cite{M3DM} & 98.4 & 98.6 & 98.9 \\
CFM~\cite{CFM} & \underline{98.6} & 98.7 & \underline{99.1} \\
G$^2$SF~\cite{g2sf} & \underline{98.6} & \underline{99.0} & \underline{99.1} \\
\midrule
Ours & \textbf{98.8} & \textbf{99.1} & \textbf{99.4} \\
\bottomrule
\end{tabular}
\end{minipage}
\end{table}

\paratitle{Pixel-level Area Under the Receiver Operating Characteristic (P-AUROC).} Tab.~\ref{supp:tab:pauroc_modality} reports P-AUROC scores for each modality on MVTec 3D-AD~\cite{mvtec3d} under the full-shot setting. Our method achieves 98.8 for RGB, 97.5 for 3D, and 99.6 for multi-modal fusion.
Although P-AUROC scores are saturated above 99.0 in the full-shot multi-modal setting, our method demonstrates competitive performance on this dataset~\cite{mvtec3d}. Tab.~\ref{supp:tab:pauroc_fewshot} further reports P-AUROC under few-shot settings (5, 10, 50 shots). As shown in the main paper (Tab.~2), our method outperforms all baselines across all metrics (e.g., I-AUROC, AUPRO@30\%, AUPRO@1\%) under the few-shot setting, and P-AUROC results under few-shot settings also align with this trend.

\begin{table}[t]
\centering
\caption{Multi-modal results on Eyecandies dataset~\cite{eyecandies}. We report I-AUROC, P-AUROC, and AUPRO metrics averaged over all 10 categories.}
\label{supp:tab:eyecandies}
\scriptsize
\setlength{\tabcolsep}{4pt}
\begin{tabular}{@{}l|cccc@{}}
\toprule
Method & I-AUROC & P-AUROC & AUPRO@30\% & AUPRO@1\% \\
\midrule
M3DM~\cite{M3DM} & 89.7 & 97.7 & 88.2 & 33.1 \\
CFM~\cite{CFM} & 88.1 & 97.3 & 88.7 & 33.5 \\
LSFA~\cite{LSFA} & \underline{90.6} & 97.4 & - & - \\
G$^2$SF~\cite{g2sf} & 90.2 & \textbf{98.2} & \underline{89.8} & \underline{35.7} \\
\midrule
Ours & \textbf{91.2} & \underline{97.8} & \textbf{90.1} & \textbf{37.6} \\
\bottomrule
\end{tabular}
\end{table}

\paratitle{Performance on Eyecandies Dataset.} We previously reported P-AUROC results on MVTec 3D-AD~\cite{mvtec3d}, and here we present multi-modal fusion results including all metrics on Eyecandies dataset~\cite{eyecandies}. As shown in Tab.~\ref{supp:tab:eyecandies}, our method achieves 91.2 in I-AUROC, 97.8 in P-AUROC, 90.1 in AUPRO@30\%, and 37.6 in AUPRO@1\%. These results demonstrate that our approach performs effectively beyond MVTec 3D-AD~\cite{mvtec3d}.

\paratitle{Generalization to Uni-modal Benchmarks.}
While EPAR targets the multi-modal setting, both feature adaptation (Ada.) and the noise suppression network (NSN) operate independently within each modality. To verify that our framework generalizes beyond the multi-modal regime, we evaluate it on two uni-modal benchmarks: RGB-only MVTec-AD~\cite{mvtecad} and 3D-only Real3D-AD~\cite{real3d}.

\begin{table}[t]
\centering
\caption{Generalization to the RGB-only MVTec-AD~\cite{mvtecad} benchmark. Starting from the frozen DINO~\cite{dino} encoder, we progressively add feature adaptation (Ada.) and the noise suppression network (NSN).}
\label{supp:tab:mvtecad_rgb}
\scriptsize
\setlength{\tabcolsep}{6pt}
\begin{tabular}{@{}l|ccc@{}}
\toprule
Configuration & I-AUROC & AUPRO@30\% & AUPRO@1\% \\
\midrule
DINO & 99.0 & 94.6 & 35.2 \\
\;+\,Ada. & 99.1 & \underline{94.6} & \underline{36.8} \\
\;+\,Ada.\,\&\,NSN & 99.1 & \textbf{94.9} & \textbf{37.1} \\
\bottomrule
\end{tabular}
\end{table}

\begin{table}[t]
\centering
\caption{Generalization to the 3D-only Real3D-AD~\cite{real3d} benchmark. We report I-AUROC for feature adaptation alone (without NSN) to isolate the impact of our adaptation strategy against contrastive adaptation.}
\label{supp:tab:real3d}
\scriptsize
\setlength{\tabcolsep}{6pt}
\begin{tabular}{@{}l|c@{}}
\toprule
Method & I-AUROC \\
\midrule
PointMAE~\cite{pointmae}\,+\,PatchCore~\cite{patchcore} & 51.1 \\
LSFA~\cite{LSFA} (contrastive adaptation) & 58.5 \\
Ours (Ada., w/o NSN) & \textbf{59.3} \\
\bottomrule
\end{tabular}
\end{table}

Tab.~\ref{supp:tab:mvtecad_rgb} reports results on the RGB-only MVTec-AD~\cite{mvtecad} benchmark, starting from the frozen DINO~\cite{dino} encoder and progressively adding feature adaptation (Ada.) and the noise suppression network (NSN). Feature adaptation alone (+Ada.) improves the segmentation metrics, and the full pipeline (+Ada.\,\&\,NSN) yields the best results across both AUPRO metrics.

Tab.~\ref{supp:tab:real3d} reports I-AUROC on the 3D-only Real3D-AD~\cite{real3d} benchmark. To isolate the effect of our adaptation strategy, we evaluate feature adaptation alone, without NSN. Our embedding prediction with patch-level VIC regularization achieves 59.3 I-AUROC, surpassing PointMAE~\cite{pointmae}+PatchCore~\cite{patchcore} (51.1, +8.2\%) and the contrastive adaptation of LSFA~\cite{LSFA} (58.5, +0.8\%). This demonstrates that our adaptation strategy outperforms contrastive adaptation even in the 3D-only setting.

\begin{table*}[t]
\centering
\caption{Impact of refinement stage across modalities. Ada.: feature adaptation only. w/ NSN: full pipeline with NSN-based refinement.}
\label{supp:tab:nsn_ablation}
\scriptsize
\begin{minipage}{0.46\textwidth}
\centering
\subcaption{RGB Modality Results}
\resizebox{\linewidth}{!}{%
\begin{tabular}{@{}l|cc|cc|cc@{}}
\toprule
& \multicolumn{2}{c|}{I-AUROC} & \multicolumn{2}{c|}{AUPRO@30\%} & \multicolumn{2}{c}{AUPRO@1\%} \\
\cmidrule(lr){2-3} \cmidrule(lr){4-5} \cmidrule(lr){6-7}
Class & Ada. & w/ NSN & Ada. & w/ NSN & Ada. & w/ NSN \\
\midrule
Bagel & 96.1 & \textbf{97.5} & 95.6 & \textbf{96.6} & 39.9 & \textbf{46.9} \\
Cable Gland & \textbf{93.7} & 93.6 & 95.2 & \textbf{96.0} & 40.3 & \textbf{43.0} \\
Carrot & 82.2 & \textbf{82.7} & 96.8 & \textbf{96.9} & 39.9 & \textbf{40.3} \\
Cookie & 80.5 & \textbf{85.9} & 93.1 & \textbf{94.5} & 32.3 & \textbf{37.5} \\
Dowel & \textbf{94.2} & 94.1 & 94.9 & \textbf{95.0} & 38.8 & \textbf{39.1} \\
Foam & 81.9 & \textbf{82.1} & 86.6 & \textbf{86.8} & \textbf{29.6} & 29.5 \\
Peach & 86.5 & \textbf{88.6} & 97.0 & \textbf{97.3} & 38.2 & \textbf{40.7} \\
Potato & 61.5 & \textbf{64.6} & 96.1 & \textbf{96.2} & 32.9 & \textbf{33.6} \\
Rope & 95.7 & \textbf{96.6} & 97.3 & \textbf{97.4} & 41.4 & \textbf{43.0} \\
Tire & \textbf{76.1} & 75.8 & \textbf{96.9} & 96.9 & 40.1 & 40.1 \\
\midrule
Mean & 84.8 & \textbf{86.2} & 95.0 & \textbf{95.4} & 37.3 & \textbf{39.4} \\
\bottomrule
\end{tabular}%
}
\end{minipage}%
\hfill
\begin{minipage}{0.46\textwidth}
\centering
\subcaption{3D Modality Results}
\resizebox{\linewidth}{!}{%
\begin{tabular}{@{}l|cc|cc|cc@{}}
\toprule
& \multicolumn{2}{c|}{I-AUROC} & \multicolumn{2}{c|}{AUPRO@30\%} & \multicolumn{2}{c}{AUPRO@1\%} \\
\cmidrule(lr){2-3} \cmidrule(lr){4-5} \cmidrule(lr){6-7}
Class & Ada. & w/ NSN & Ada. & w/ NSN & Ada. & w/ NSN \\
\midrule
Bagel & 97.9 & \textbf{98.6} & 97.3 & 97.3 & 47.1 & \textbf{47.9} \\
Cable Gland & \textbf{83.6} & 82.9 & 94.0 & 94.0 & \textbf{33.8} & 33.0 \\
Carrot & 98.8 & \textbf{98.9} & 98.1 & 98.1 & 48.2 & \textbf{48.6} \\
Cookie & \textbf{99.5} & 99.2 & 94.8 & \textbf{94.9} & 47.0 & \textbf{48.2} \\
Dowel & \textbf{95.3} & 95.0 & 82.8 & \textbf{83.0} & 33.3 & \textbf{33.4} \\
Foam & 83.6 & \textbf{83.7} & 84.2 & \textbf{84.4} & 31.4 & 31.4 \\
Peach & 96.9 & \textbf{99.2} & 98.2 & \textbf{98.3} & 49.2 & \textbf{51.1} \\
Potato & 96.0 & \textbf{97.1} & 98.3 & \textbf{98.4} & 49.6 & \textbf{50.6} \\
Rope & 98.6 & 98.6 & 95.9 & 95.9 & 44.3 & \textbf{44.7} \\
Tire & 99.1 & 99.1 & 98.2 & 98.2 & 48.7 & \textbf{48.8} \\
\midrule
Mean & 94.9 & \textbf{95.2} & 94.2 & \textbf{94.3} & 43.3 & \textbf{43.8} \\
\bottomrule
\end{tabular}%
}
\end{minipage}
\end{table*}

\paratitle{Impact of Refinement Stage across Modalities.} Tab.~\ref{supp:tab:nsn_ablation} presents the contribution of each stage in our framework: feature adaptation alone (Ada.) versus the full pipeline with NSN-based refinement (w/ NSN). For RGB, the refinement stage improves I-AUROC from 84.8 to 86.2 (+1.4\%), AUPRO@30\% from 95.0 to 95.4 (+0.4\%), and AUPRO@1\% from 37.3 to 39.4 (+2.1\%). For 3D, we observe consistent improvements with I-AUROC from 94.9 to 95.2 (+0.3\%), AUPRO@30\% from 94.2 to 94.3 (+0.1\%), and AUPRO@1\% from 43.3 to 43.8 (+0.5\%). These results demonstrate that our NSN-based refinement stage provides further improvements building upon feature adaptation.

\begin{table}[t]
\centering
\caption{Per-modality comparison of refinement strategies in AUPRO@1\% on MVTec 3D-AD. Direct directly refines the coarse pixel-level map following ReConPatch~\cite{reconpatch}, whereas Modulate (ours) uses the NSN refinement map to modulate the coarse map (Eq.~8 of the main paper).}
\label{supp:tab:direct_refine}
\scriptsize
\setlength{\tabcolsep}{8pt}
\begin{tabular}{@{}l|ccc@{}}
\toprule
Strategy & RGB & 3D & Fusion \\
\midrule
Direct~\cite{reconpatch} & 25.0 & 30.0 & 34.1 \\
Modulate (Ours) & \textbf{39.4} & \textbf{43.8} & \textbf{48.6} \\
\bottomrule
\end{tabular}
\end{table}

\paratitle{Per-modality Analysis of Refinement Strategies.} We further analyze the per-modality behavior of the two refinement strategies compared in the main paper (Tab.~6): Direct, which directly refines the coarse pixel-level map following ReConPatch~\cite{reconpatch}, and Modulate (ours), which uses the NSN refinement map to modulate the coarse map (Eq.~8 of the main paper). Tab.~\ref{supp:tab:direct_refine} reports AUPRO@1\% for each modality. Direct refinement underperforms across all modalities (RGB 25.0, 3D 30.0, Fusion 34.1), even falling below the coarse map without refinement (47.6, Tab.~5 of the main paper). This is because directly fitting the pixel-level map risks learning the appearance of synthetic anomalies, which differs explicitly from real defects and thus fails to generalize at inference. In contrast, our Modulate strategy preserves the coarse distance-based map and selectively amplifies true anomaly signals, achieving substantially higher AUPRO@1\% across all modalities (RGB 39.4, 3D 43.8, Fusion 48.6). This confirms that learning the relational feature discrepancy, rather than the synthetic anomaly appearance, is the key to generalizable refinement.

\begin{table}[t]
\centering
\caption{Comparison of pseudo-anomaly synthesis strategies on MVTec 3D-AD (RGB-only).}
\label{supp:tab:supp_synthesis_rgb}
\scriptsize
\setlength{\tabcolsep}{4pt}
\begin{tabular}{lcccc}
\toprule
Synthesis & I-AUROC & P-AUROC & AUPRO@30\% & AUPRO@1\% \\
\midrule
Within-dataset & 86.8 & 98.9 & 95.2 & 39.2 \\
DTD            & 86.2 & 98.8 & 95.4 & 39.4 \\
\bottomrule
\end{tabular}
\end{table}

\paratitle{Robustness to the Pseudo-Anomaly Synthesis Pipeline (RGB-only).}
In the main paper (Tab.~6~(b)), we show that NSN is robust to the pseudo-anomaly synthesis strategy. Here, we report the RGB-only results, since the choice between external (DTD~\cite{dtd}) and within-dataset synthesis is intrinsic to the RGB modality. As shown in Tab.~\ref{supp:tab:supp_synthesis_rgb}, the two strategies yield nearly identical results across all metrics, confirming that NSN is not sensitive to the synthesis strategy.

\begin{table}[t]
\centering
\caption{Ablation study on the VIC regularization coefficient ($\alpha$ in Eq.~3 of the main paper) under the multi-scale configuration, evaluated with feature adaptation only (Ada.) and with the full pipeline (Ada.\,+\,NSN). Results are reported on MVTec 3D-AD.}
\label{supp:tab:vic_alpha}
\scriptsize
\setlength{\tabcolsep}{4pt}
\begin{tabular}{@{}c|cc|ccc@{}}
\toprule
$\alpha$ & Ada. & NSN & I-AUROC & AUPRO@30\% & AUPRO@1\% \\
\midrule
\multirow{2}{*}{0.0}
& \checkmark & & 95.7 & 97.6 & 47.5 \\
& \checkmark & \checkmark & 97.0 & 97.6 & 48.2 \\
\midrule
\multirow{2}{*}{0.001}
& \checkmark & & 97.7 & 97.6 & 47.7 \\
& \checkmark & \checkmark & 98.2 & 97.7 & 48.5 \\
\midrule
\multirow{2}{*}{0.01}
& \checkmark & & 98.1 & 97.7 & 47.6 \\
& \checkmark & \checkmark & \textbf{98.4} & \textbf{97.8} & \textbf{48.6} \\
\bottomrule
\end{tabular}
\end{table}

\paratitle{Hyperparameter of the VIC Regularization Coefficient.} We analyze the sensitivity of our framework to the VIC regularization coefficient $\alpha$, which balances the embedding prediction loss and the VIC regularization term (Eq.~3 in the main paper). For the VIC regularization, we set $\lambda_{\text{var}} = 15$, $\lambda_{\text{inv}} = 15$, and $\lambda_{\text{cov}} = 1$. Tab.~\ref{supp:tab:vic_alpha} reports results under the multi-scale configuration for both feature adaptation alone (Ada.) and the full pipeline (Ada.\,+\,NSN). Without VIC regularization ($\alpha=0.0$), feature adaptation attains competitive segmentation but suboptimal sample-level detection (95.7 I-AUROC). Introducing VIC regularization with a small coefficient ($\alpha=0.001$) already recovers most of the gain (97.7 I-AUROC), and $\alpha=0.01$ yields the best results across all metrics. We therefore adopt $\alpha=0.01$ in all main experiments both MVTec 3D-AD~\cite{mvtec3d} and Eyecandies~\cite{eyecandies}.

\begin{table}[t]
\centering\footnotesize
\setlength{\tabcolsep}{2pt}
\renewcommand{\arraystretch}{0.9}
\caption{Fusion strategy ablation on MVTec 3D-AD.
PRO@30\% and PRO@1\% denote AUPRO at FPR thresholds of 30\% and 1\%, respectively.}
\label{supp:tab:fusion_comp}
\begin{subtable}[t]{0.50\linewidth}\centering
\caption{Map fusion (segmentation)}
\begin{tabular}{lcc}
\toprule
Fusion & PRO@30\% & PRO@1\% \\
\midrule
Mean & 96.9 & 45.7 \\
Hadamard & 97.4 & 48.3 \\
\textbf{Max} & \textbf{97.8} & \textbf{48.6} \\
\bottomrule
\end{tabular}
\end{subtable}
\hfill
\begin{subtable}[t]{0.47\linewidth}\centering
\caption{Score aggregation (detection)}
\begin{tabular}{lc}
\toprule
Aggregation & I-AUROC \\
\midrule
Sum & 87.3 \\
Maximum & 97.5 \\
\textbf{Product} & \textbf{98.4} \\
\bottomrule
\end{tabular}
\end{subtable}
\end{table}

\noindent\textbf{Fusion Strategy.} We analyze our late-fusion strategy for combining per-modality outputs, separately for segmentation and detection (Tab.~\ref{supp:tab:fusion_comp}).
(a) For pixel-level map fusion, we compare mean, Hadamard product~\cite{hadamard}, and element-wise maximum after z-score normalization. We observe that maximum fusion performs best (Tab.~\ref{supp:tab:fusion_comp}~(a)), preserving an anomalous response whenever either modality detects it.
(b) For sample-level score aggregation, we compare sum, maximum, and product. Product performs best, surpassing maximum by +0.9\% in I-AUROC (Tab.~\ref{supp:tab:fusion_comp}~(b)). 
While maximum, acting as an OR-gate, remains competitive (0.975 vs.\ 0.971 for G$^2$SF~\cite{g2sf}), product acts as an AND-gate that requires cross-modal agreement, suppressing single-modality false positives. Such AND-gate behavior is more beneficial for detection than propagating every modality-specific activation.

\begin{table}[t]
\centering
\caption{Peak GPU memory consumption at inference on MVTec 3D-AD. We use G$^2$SF~\cite{g2sf} as the reference and report relative memory under identical conditions.}
\label{supp:tab:mem_overhead}
\scriptsize
\setlength{\tabcolsep}{6pt}
\begin{tabular}{@{}l|cc@{}}
\toprule
Method & Peak GPU Mem. & Relative \\
\midrule
G$^2$SF~\cite{g2sf} & 3404 MB & 1.00$\times$ \\
Ours (Multi, 10\%) & 6183 MB & 1.83$\times$ \\
Ours (Single, 5\% coreset) & \textbf{2828 MB} & \textbf{0.83$\times$} \\
\bottomrule
\end{tabular}
\end{table}

\paratitle{Memory Overhead.} Beyond the inference latency analysis in the main paper (Tab.~7), we further report peak GPU memory consumption at inference, another key constraint for industrial deployment. As shown in Tab.~\ref{supp:tab:mem_overhead}, using G$^2$SF~\cite{g2sf} as the reference (3404 MB, 1.00$\times$), our multi-scale configuration with a 10\% memory bank consumes 6183 MB (1.83$\times$), primarily due to processing high-dimensional multi-layer features. In contrast, our single-scale configuration with a 5\% coreset requires only 2828 MB (0.83$\times$)---less memory than G$^2$SF---while still outperforming it in I-AUROC and AUPRO@1\% (Tab.~7 of the main paper). This confirms that EPAR offers a favorable accuracy--efficiency trade-off even under tight memory budgets.

%%%%%%%%%%%%%%%%%%%%%%%%%%%%%%%%%%%%%%%%%%%%%%%%%%%%%%%%%%%%%
% SECTION D: Qualitative Results
%%%%%%%%%%%%%%%%%%%%%%%%%%%%%%%%%%%%%%%%%%%%%%%%%%%%%%%%%%%%%
\section{Qualitative Results}
\label{supp:sec:qualitative_res}

\vspace{-5em}

\begin{figure}[H]
\centering
\includegraphics[width=\linewidth]{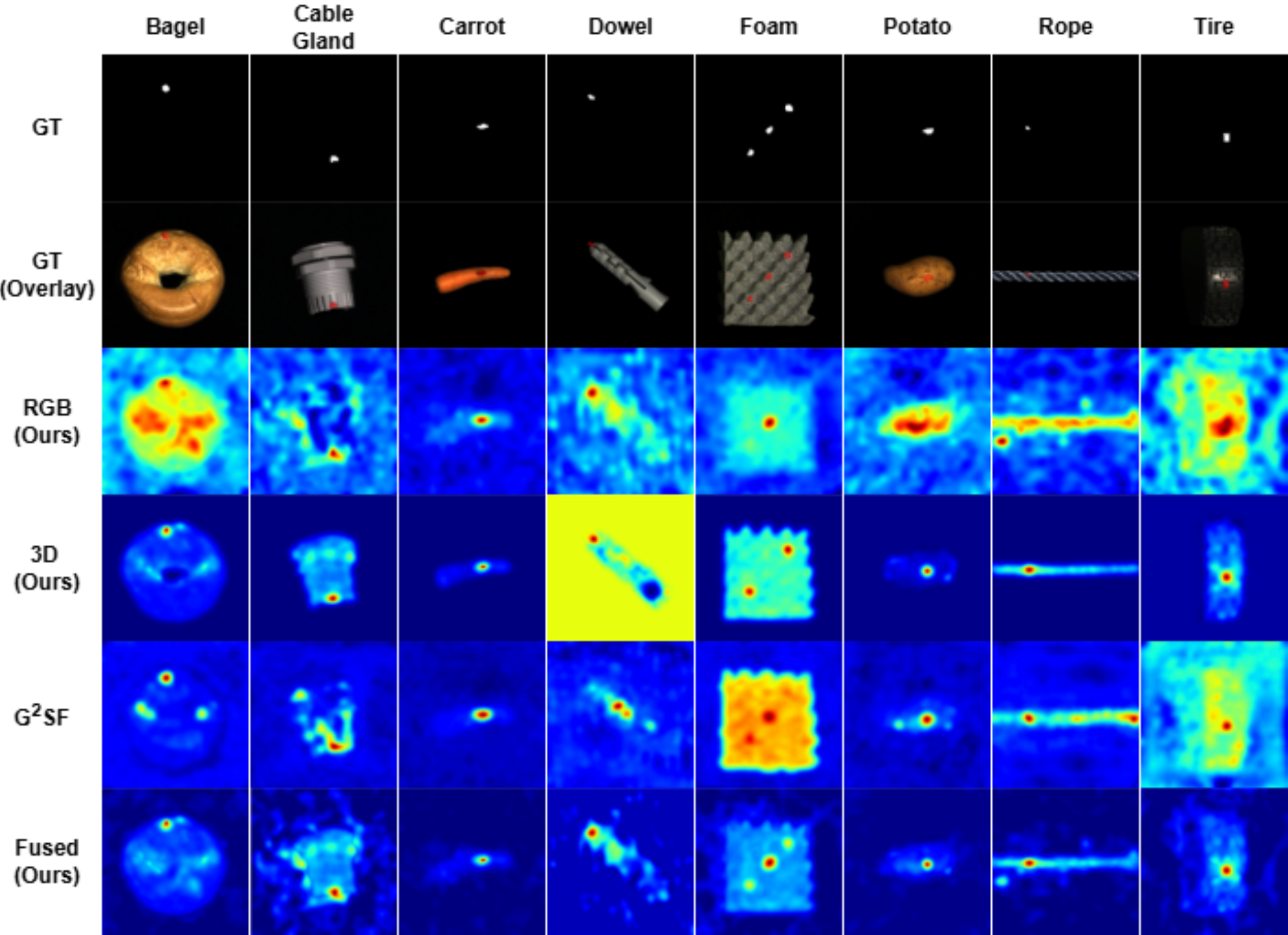}
\caption{Qualitative results on MVTec 3D-AD dataset. GT (Overlay) shows the ground truth anomaly regions overlaid on the RGB image.}
\label{supp:fig:mvtec_qual}
\end{figure}

\vspace{-1.2em}
\begin{figure}[H]
\centering
\includegraphics[width=\linewidth]{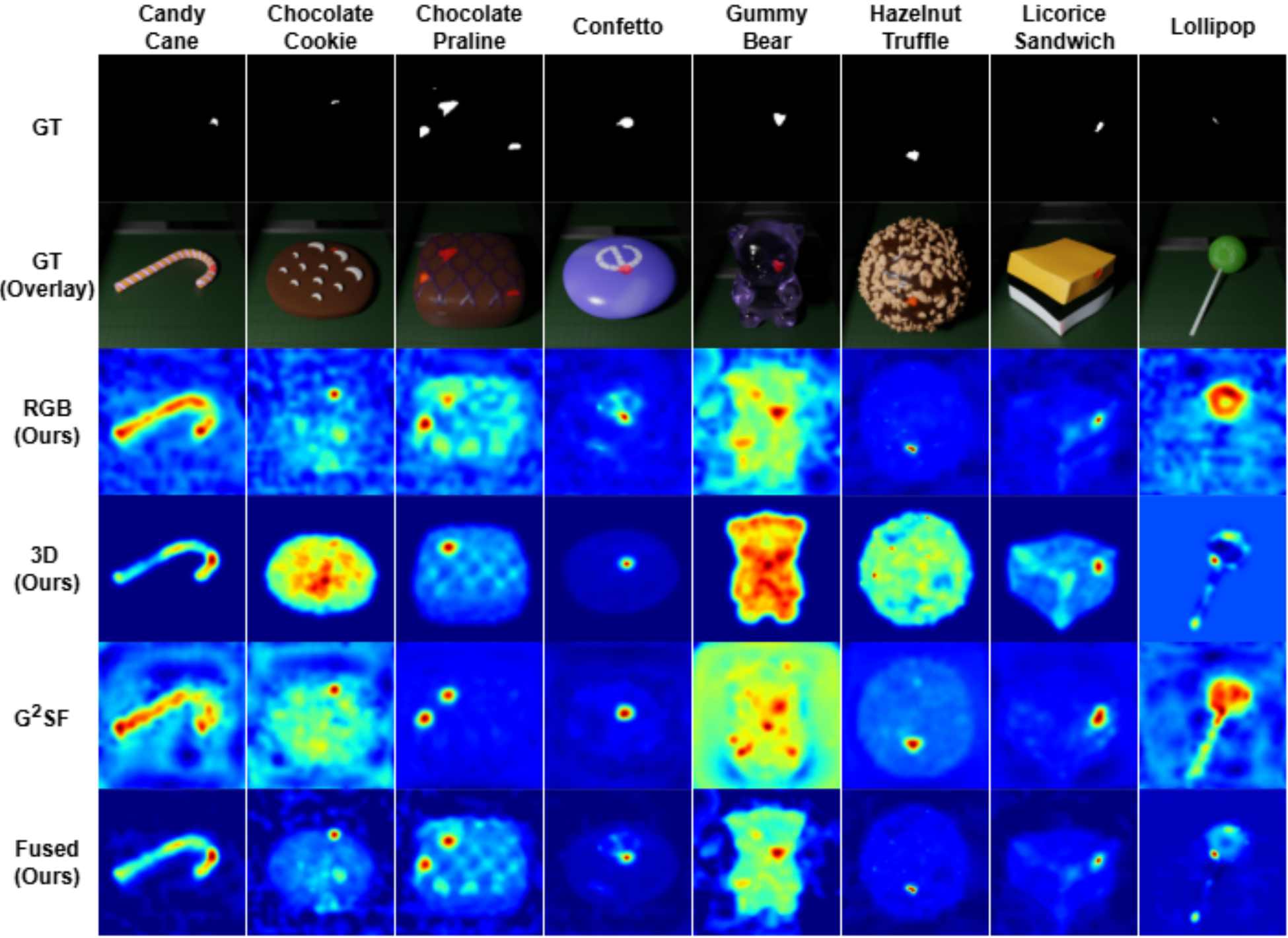}
\caption{Qualitative results on Eyecandies dataset. GT (Overlay) shows the ground truth anomaly regions overlaid on the RGB image.}
\label{supp:fig:eyecandies_qual}
\end{figure}

%%%%%%%%%%%%%%%%%%%%%%%%%%%%%%%%%%%%%%%%%%%%%%%%%%%%%%%%%%%%%
% SECTION E: Detailed Feature Analysis
%%%%%%%%%%%%%%%%%%%%%%%%%%%%%%%%%%%%%%%%%%%%%%%%%%%%%%%%%%%%%
\section{Detailed Feature Analysis on MVTec 3D-AD}
\label{supp:sec:feature_analysis}

In this section, we provide two in-depth empirical analyses on the patch-level features extracted from normal samples. (i) In Sec.~4.4 of the main paper, we argued that our adaptation strategy encodes spatial dependencies for precise anomaly segmentation, unlike contrastive-based adaptation~\cite{LSFA}. To examine how spatial relationships are captured, we conduct cosine similarity-based feature analysis for both RGB and 3D modalities. (ii) In Sec.~4.5 of the main paper, we compare batch-level and patch-level VIC regularization. Here, we extend this analysis by providing per-category breakdown to examine how these feature statistics change across diverse object categories.

\subsection{Analysis of Feature Adaptation on Spatial Relationship}
\label{supp:subsec:spatial_analysis}

We analyze how our adaptation strategy affects spatial relationships in feature space through both qualitative visualization and quantitative metrics.

\begin{figure}[t]
    \centering
    \includegraphics[width=1.0\linewidth]{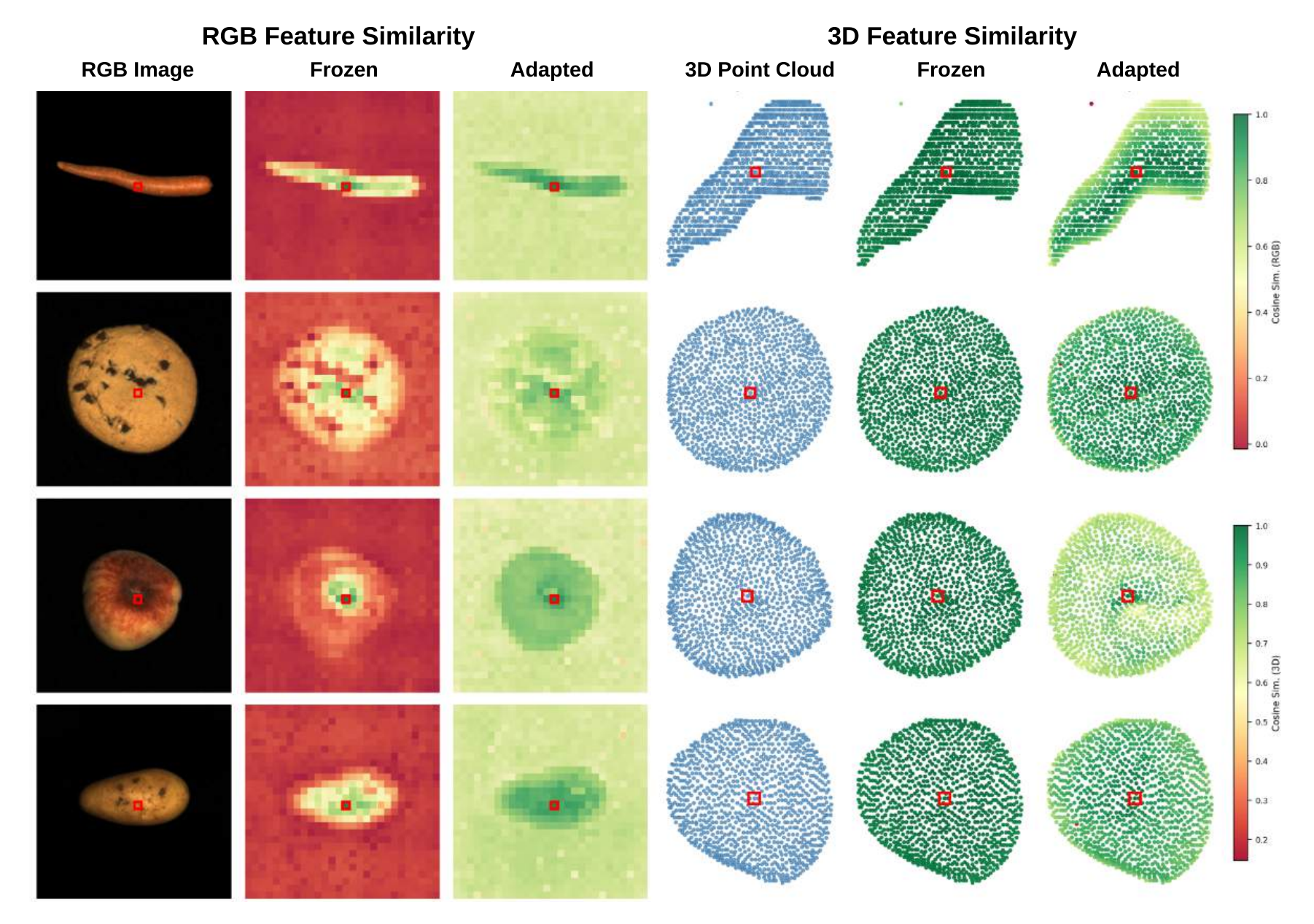}
    \caption{Visualization of feature similarity before and after adaptation. We compute cosine similarity between a query feature (red marker) and all other features. For RGB, adapted features exhibit smooth spatial transitions compared to frozen features. For 3D, adapted features show clear local-distant contrast, whereas frozen features display nearly uniform similarity across all points.}
    \label{supp:fig:spatial_analysis}
\end{figure}

\paratitle{Qualitative Analysis.} Fig.~\ref{supp:fig:spatial_analysis} visualizes cosine similarity between a query feature (red marker) and all other features, comparing frozen pretrained features and adapted features. For RGB, frozen features exhibit irregular similarity patterns with abrupt changes even between spatially adjacent patches. After adaptation, features show smooth spatial transitions where similarity gradually decreases as distance from the query increases. For 3D, frozen features exhibit nearly uniform similarity across all points, indicating a lack of spatial differentiation. In contrast, adapted features capture local spatial relationships, showing high similarity for nearby points and progressively lower similarity for distant points.

\begin{table}[t]
\centering
\caption{Spatial Relationship Encoding Metrics. We measure Neighborhood Similarity (NS) for RGB and Local-Distant Contrast (LDC) for 3D point clouds across all categories.}
\label{supp:tab:spatial_metrics}
\begin{subtable}[t]{\linewidth}
\centering
\scriptsize
\caption{Neighborhood Similarity (NS) - RGB}
\label{supp:tab:ns_rgb}
\begin{tabular}{lccc}
\toprule
Class & Frozen & Adapted & $\Delta$ \\
\midrule
Bagel & 0.8068 & 0.8952 & +0.0885 \\
Cable Gland & 0.8293 & 0.9245 & +0.0952 \\
Carrot & 0.8540 & 0.9301 & +0.0761 \\
Cookie & 0.7454 & 0.8579 & +0.1125 \\
Dowel & 0.8182 & 0.9016 & +0.0833 \\
Foam & 0.7282 & 0.8680 & +0.1398 \\
Peach & 0.8266 & 0.9036 & +0.0770 \\
Potato & 0.8590 & 0.9357 & +0.0767 \\
Rope & 0.9179 & 0.9560 & +0.0382 \\
Tire & 0.8234 & 0.9293 & +0.1059 \\
\midrule
\textbf{Mean} & \textbf{0.8209} & \textbf{0.9102} & \textbf{+0.0893} \\
\bottomrule
\end{tabular}
\end{subtable}

\vspace{1em}

\begin{subtable}[t]{\linewidth}
\centering
\scriptsize
\caption{Local-Distant Contrast (LDC) - 3D}
\label{supp:tab:ldc_3d}
\begin{tabular}{lccc}
\toprule
Class & Frozen & Adapted & $\Delta$ \\
\midrule
Bagel & 0.1075 & 0.2296 & +0.1221 \\
Cable Gland & 0.0574 & 0.1869 & +0.1295 \\
Carrot & 0.0612 & 0.2542 & +0.1930 \\
Cookie & 0.0754 & 0.1932 & +0.1178 \\
Dowel & 0.0496 & 0.1774 & +0.1278 \\
Foam & 0.1357 & 0.0694 & -0.0662 \\
Peach & 0.0716 & 0.2527 & +0.1811 \\
Potato & 0.0669 & 0.2466 & +0.1797 \\
Rope & 0.0789 & 0.1733 & +0.0944 \\
Tire & 0.1067 & 0.1261 & +0.0193 \\
\midrule
\textbf{Mean} & \textbf{0.0811} & \textbf{0.1909} & \textbf{+0.1099} \\
\bottomrule
\end{tabular}
\end{subtable}
\end{table}

\begin{table}[t]
\centering
\caption{Effect of feature adaptation on segmentation performance for RGB (top) and 3D (bottom) modalities.}
\label{supp:tab:feature_adaptation}
\begin{minipage}{\linewidth}
\centering
\subcaption{RGB Modality Results}
\scriptsize
\begin{tabular}{@{}l|ccc|ccc@{}}
\toprule
& \multicolumn{3}{c|}{AUPRO@30\%} & \multicolumn{3}{c}{AUPRO@1\%} \\
\cmidrule(lr){2-4} \cmidrule(lr){5-7}
Class & Frozen & Adapted & $\Delta$ & Frozen & Adapted & $\Delta$ \\
\midrule
Bagel & 94.8 & 95.6 & +0.8 & 35.8 & 39.9 & +4.1 \\
Cable Gland & 97.4 & 95.2 & -2.2 & 43.2 & 40.3 & -2.9 \\
Carrot & 97.5 & 96.8 & -0.7 & 43.1 & 39.9 & -3.2 \\
Cookie & 87.2 & 93.1 & +5.9 & 23.4 & 32.3 & +8.9 \\
Dowel & 94.9 & 94.9 & +0.0 & 38.4 & 38.8 & +0.4 \\
Foam & 88.4 & 86.6 & -1.8 & 31.3 & 29.6 & -1.7 \\
Peach & 97.3 & 97.0 & -0.3 & 41.7 & 38.2 & -3.5 \\
Potato & 95.5 & 96.1 & +0.6 & 27.5 & 32.9 & +5.4 \\
Rope & 97.2 & 97.3 & +0.1 & 42.6 & 41.4 & -1.2 \\
Tire & 96.6 & 96.9 & +0.3 & 39.9 & 40.1 & +0.2 \\
\midrule
Mean & 94.7 & 95.0 & +0.3 & 36.7 & 37.3 & +0.6 \\
\bottomrule
\end{tabular}
\end{minipage}

% \vspace{1em}

\begin{minipage}{\linewidth}
\centering
\subcaption{3D Modality Results}
\scriptsize
\begin{tabular}{@{}l|ccc|ccc@{}}
\toprule
& \multicolumn{3}{c|}{AUPRO@30\%} & \multicolumn{3}{c}{AUPRO@1\%} \\
\cmidrule(lr){2-4} \cmidrule(lr){5-7}
Class & Frozen & Adapted & $\Delta$ & Frozen & Adapted & $\Delta$ \\
\midrule
Bagel & 83.6 & 97.3 & +13.7 & 21.5 & 47.1 & +25.6 \\
Cable Gland & 60.2 & 94.0 & +33.8 & 3.1 & 33.8 & +30.7 \\
Carrot & 87.3 & 98.1 & +10.8 & 7.4 & 48.2 & +40.8 \\
Cookie & 94.2 & 94.8 & +0.6 & 39.7 & 47.0 & +7.3 \\
Dowel & 0.0 & 82.8 & +82.8 & 0.0 & 33.3 & +33.3 \\
Foam & 56.8 & 84.2 & +27.4 & 4.3 & 31.4 & +27.1 \\
Peach & 84.9 & 98.2 & +13.3 & 17.3 & 49.2 & +31.9 \\
Potato & 0.0 & 98.3 & +98.3 & 0.0 & 49.6 & +49.6 \\
Rope & 91.4 & 95.9 & +4.5 & 25.5 & 44.3 & +18.8 \\
Tire & 63.4 & 98.2 & +34.8 & 3.5 & 48.7 & +45.2 \\
\midrule
Mean & 62.2 & 94.2 & +32.0 & 12.2 & 43.3 & +31.1 \\
\bottomrule
\end{tabular}
\end{minipage}
\end{table}

\begin{table}[t]
\centering
\caption{Per-category feature analysis of VIC regularization on MVTec 3D-AD. (a) Effect of variance regularization: w/ VIC vs w/o VIC (per-category). (b) Comparison between VIC and covariance-only (COV) regularization (per-category). (c) Aggregated feature statistics and their correlation with detection and segmentation performance. (d) Aggregated Feature statistics under batch-level and patch-level VIC regularization.}
\label{supp:tab:vic_anal}
\footnotesize
\begin{minipage}{\linewidth}
\centering
\subcaption{w/ VIC vs w/o VIC}
\setlength{\tabcolsep}{2pt}
\begin{tabular}{@{}l|cc|cc|cc|cc@{}}
\toprule
& \multicolumn{4}{c|}{RGB} & \multicolumn{4}{c}{3D} \\
\cmidrule(lr){2-5} \cmidrule(lr){6-9}
& \multicolumn{2}{c|}{Variance $\uparrow$} & \multicolumn{2}{c|}{Covariance $\downarrow$} & \multicolumn{2}{c|}{Variance $\uparrow$} & \multicolumn{2}{c}{Covariance $\downarrow$} \\
Class & VIC & w/o & VIC & w/o & VIC & w/o & VIC & w/o \\
\midrule
Bagel & \textbf{0.490} & 0.482 & \textbf{0.052} & 0.053 & \textbf{0.303} & 0.291 & \textbf{0.039} & 0.067 \\
Cable G. & 0.406 & \textbf{0.593} & \textbf{0.032} & 0.082 & 0.206 & \textbf{0.225} & \textbf{0.015} & 0.026 \\
Carrot & 0.371 & \textbf{0.539} & \textbf{0.034} & 0.066 & 0.124 & \textbf{0.133} & \textbf{0.009} & 0.016 \\
Cookie & 0.527 & \textbf{0.530} & \textbf{0.051} & 0.060 & 0.276 & \textbf{0.284} & \textbf{0.053} & 0.058 \\
Dowel & 0.409 & \textbf{0.550} & \textbf{0.043} & 0.072 & 0.138 & \textbf{0.140} & \textbf{0.014} & 0.015 \\
Foam & 0.508 & \textbf{0.567} & \textbf{0.049} & 0.076 & \textbf{0.266} & 0.260 & \textbf{0.020} & 0.056 \\
Peach & 0.452 & \textbf{0.613} & \textbf{0.043} & 0.086 & \textbf{0.246} & 0.241 & \textbf{0.026} & 0.050 \\
Potato & 0.389 & \textbf{0.541} & \textbf{0.032} & 0.061 & 0.164 & \textbf{0.167} & \textbf{0.017} & 0.024 \\
Rope & 0.399 & \textbf{0.478} & \textbf{0.057} & 0.070 & 0.113 & \textbf{0.121} & \textbf{0.006} & 0.013 \\
Tire & 0.448 & \textbf{0.636} & \textbf{0.047} & 0.088 & 0.189 & \textbf{0.199} & \textbf{0.017} & 0.032 \\
\midrule
Mean & 0.440 & \textbf{0.553} & \textbf{0.044} & 0.071 & 0.203 & \textbf{0.206} & \textbf{0.022} & 0.036 \\
\bottomrule
\end{tabular}
\end{minipage}

\vspace{2em}

\begin{minipage}{\linewidth}
\centering
\subcaption{VIC vs COV only}
\setlength{\tabcolsep}{2pt}
\begin{tabular}{@{}l|cc|cc|cc|cc@{}}
\toprule
& \multicolumn{4}{c|}{RGB} & \multicolumn{4}{c}{3D} \\
\cmidrule(lr){2-5} \cmidrule(lr){6-9}
& \multicolumn{2}{c|}{Variance $\uparrow$} & \multicolumn{2}{c|}{Covariance $\downarrow$} & \multicolumn{2}{c|}{Variance $\uparrow$} & \multicolumn{2}{c}{Covariance $\downarrow$} \\
Class & VIC & COV & VIC & COV & VIC & COV & VIC & COV \\
\midrule
Bagel & \textbf{0.490} & 0.177 & 0.052 & \textbf{0.008} & \textbf{0.303} & 0.268 & \textbf{0.039} & 0.068 \\
Cable G. & \textbf{0.406} & 0.168 & 0.032 & \textbf{0.006} & 0.206 & \textbf{0.210} & \textbf{0.015} & 0.044 \\
Carrot & \textbf{0.371} & 0.180 & 0.034 & \textbf{0.008} & 0.124 & \textbf{0.131} & \textbf{0.009} & 0.017 \\
Cookie & \textbf{0.527} & 0.453 & 0.051 & \textbf{0.040} & \textbf{0.276} & 0.273 & \textbf{0.053} & 0.053 \\
Dowel & \textbf{0.409} & 0.255 & 0.043 & \textbf{0.017} & \textbf{0.138} & 0.134 & \textbf{0.014} & 0.015 \\
Foam & \textbf{0.508} & 0.154 & 0.049 & \textbf{0.005} & 0.266 & \textbf{0.274} & \textbf{0.020} & 0.074 \\
Peach & \textbf{0.452} & 0.144 & 0.043 & \textbf{0.004} & \textbf{0.246} & 0.228 & \textbf{0.026} & 0.051 \\
Potato & \textbf{0.389} & 0.239 & 0.032 & \textbf{0.012} & \textbf{0.164} & 0.158 & \textbf{0.017} & 0.024 \\
Rope & \textbf{0.399} & 0.200 & 0.057 & \textbf{0.013} & 0.113 & \textbf{0.126} & \textbf{0.006} & 0.016 \\
Tire & \textbf{0.448} & 0.243 & 0.047 & \textbf{0.014} & \textbf{0.189} & 0.185 & \textbf{0.017} & 0.033 \\
\midrule
Mean & \textbf{0.440} & 0.221 & 0.044 & \textbf{0.013} & \textbf{0.203} & 0.199 & \textbf{0.022} & 0.039 \\
\bottomrule
\end{tabular}
\end{minipage}

\vspace{2.5em}

\begin{minipage}{\linewidth}
\centering
\subcaption{Feature statistics alongside performance}
\setlength{\tabcolsep}{3pt}
\begin{tabular}{@{}l|cc|cc|ccc@{}}
\toprule
& \multicolumn{2}{c|}{RGB} & \multicolumn{2}{c|}{3D} & \multicolumn{3}{c}{Performance} \\
\cmidrule(lr){2-3} \cmidrule(lr){4-5} \cmidrule(lr){6-8}
Setting & Var $\uparrow$ & Cov $\downarrow$ & Var $\uparrow$ & Cov $\downarrow$ & I-AUROC & @30\% & @1\% \\
\midrule
w/o VIC & 0.553 & 0.071 & 0.206 & 0.036 & 97.0 & 97.6 & 48.2 \\
COV only & 0.221 & 0.013 & 0.199 & 0.039 & 87.1 & 97.2 & 44.8 \\
VIC & 0.440 & 0.044 & 0.203 & 0.022 & \textbf{98.4} & \textbf{97.8} & \textbf{48.6} \\
\bottomrule
\end{tabular}
\end{minipage}

\vspace{2em}

\begin{minipage}{\linewidth}
\centering
\subcaption{Batch-level vs.\ patch-level granularity}
\setlength{\tabcolsep}{3pt}
\begin{tabular}{@{}l|cc|cc@{}}
\toprule
& \multicolumn{2}{c|}{RGB} & \multicolumn{2}{c}{3D} \\
\cmidrule(lr){2-3} \cmidrule(lr){4-5}
Granularity & Var $\uparrow$ & Cov $\downarrow$ & Var $\uparrow$ & Cov $\downarrow$ \\
\midrule
Batch-level & \textbf{0.481} & 0.050 & \textbf{0.215} & 0.030 \\
Patch-level & 0.440 & \textbf{0.044} & 0.203 & \textbf{0.022} \\
\bottomrule
\end{tabular}
\end{minipage}
\end{table}

\paratitle{Quantitative Analysis.} As shown in Fig.~\ref{supp:fig:spatial_analysis}, RGB and 3D exhibit different spatial relationship patterns, so we employ different metrics for each modality. For RGB, we measure feature similarity between each patch and its spatially neighboring patches (4 patches), defining \textbf{Neighborhood Similarity (NS)} as:
\begin{equation}
    \mathrm{NS} = \frac{1}{|\mathcal{P}|} \sum_{i \in \mathcal{P}} \frac{1}{|\mathcal{N}_i|} \sum_{j \in \mathcal{N}_i} \mathrm{sim}(f_i, f_j)
\end{equation}
where $\mathcal{P}$ denotes the set of all patches, $\mathcal{N}_i$ denotes the spatially adjacent patches of patch $i$, and $\mathrm{sim}(\cdot, \cdot)$ denotes cosine similarity. Higher NS indicates high similarity between each patch feature and its spatially adjacent patches across all patch features. For 3D, we measure local spatial relationships by computing the difference between the average similarity of each feature with its spatially nearby features and that with its spatially distant features, defining \textbf{Local-Distant Contrast (LDC)} as:
\begin{equation}
    \mathrm{LDC} = \frac{1}{|\mathcal{G}|} \sum_{i \in \mathcal{G}} \left( \frac{1}{k} \sum_{j \in \mathcal{N}_i^k} \mathrm{sim}(f_i, f_j) - \frac{1}{k} \sum_{j \in \mathcal{F}_i^k} \mathrm{sim}(f_i, f_j) \right)
\end{equation}
where $\mathcal{G}$ denotes the set of all point groups, and $\mathcal{N}_i^k$ and $\mathcal{F}_i^k$ denote the nearest and farthest $k$ groups from group $i$ in geometric distance, respectively. We set $k$=10 in our experiments. Higher LDC indicates a larger gap in similarity between nearby and distant features.

Tab.~\ref{supp:tab:spatial_metrics} and Tab.~\ref{supp:tab:feature_adaptation} report per-category results for spatial metrics and segmentation performance, respectively. For RGB, adapted features consistently show higher NS across all categories (+0.0893 on average), with modest improvements in segmentation performance. For 3D, adapted features show higher LDC in most categories (+0.1099 on average), and this enhanced local-distant differentiation leads to substantial improvements in segmentation performance. These results suggest that enhanced spatial relationship encoding is beneficial for anomaly segmentation, particularly for 3D modality where frozen pretrained features lack spatial differentiation.

\subsection{Analysis of VIC Regularization through Feature Statistics}
\label{supp:subsec:vic_analysis}

In Sec.~4.5 of the main paper, we compare batch-level and patch-level VIC regularization in terms of detection and segmentation performance. Here, we analyze the feature statistics underlying this comparison. We first provide a per-category breakdown under patch-level VIC regularization to examine how it shapes feature diversity (variance) and decorrelation (covariance) across diverse object categories, and then compare the corresponding statistics between batch-level and patch-level granularity.

\paratitle{Per-category Analysis of Feature Diversity and Decorrelation.} 
With VIC regularization applied at the patch level, we provide a per-category breakdown of the resulting patch-level feature statistics. Tab.~\ref{supp:tab:vic_anal}(a) and (b) report per-category statistics for both RGB and 3D modalities, while Tab.~\ref{supp:tab:vic_anal}(c) summarizes the aggregate statistics alongside detection and segmentation performance in the multi-modal setting.
We observe that applying full VIC regularization reduces inter-dimensional redundancy for both modalities (Tab.~\ref{supp:tab:vic_anal}(a)). This follows from the covariance term, which decorrelates feature dimensions so that each encodes distinct information. The variance term, in contrast, only maintains per-dimension variance above a target level. Its effect is thus not apparent from the full-VIC setting, where variance is no higher than without regularization. To verify that this variance floor is indeed necessary, we evaluate applying only the covariance term (COV) in Tab.~\ref{supp:tab:vic_anal}(b). Without the variance term, variance drops for both modalities. As shown in Tab.~\ref{supp:tab:vic_anal}(c), applying the full VIC regularization yields the best performance on both detection and segmentation, indicating that decorrelation alone is not sufficient. Effective anomaly detection and segmentation thus require two properties: (i) maintaining sufficient feature diversity, and (ii) reducing inter-dimensional redundancy so that each feature dimension encodes distinct information.

\paratitle{Batch-level vs.\ Patch-level Granularity.}
We then compare the feature statistics under the two granularities at which VIC regularization is applied. In Tab.~4 of the main paper, the patch-level formulation outperforms the batch-level one across all metrics. As shown in Tab.~\ref{supp:tab:vic_anal}(d), applying the regularization directly to each patch feature---rather than to its mean-pooled representation---attains lower inter-dimensional covariance in both modalities ($0.050 \to 0.044$ for RGB and $0.030 \to 0.022$ for 3D) while keeping variance comparable. The resulting patch features therefore retain their variance while carrying less redundant information across dimensions, which benefits memory bank-based IAD, where anomalies are judged by per-patch feature discrepancy.

\end{document}